\documentclass[11pt]{article}

\usepackage[final]{acl}

\usepackage{times}
\usepackage{latexsym}

\usepackage[T1]{fontenc}

\usepackage[utf8]{inputenc}

\usepackage{microtype}

\usepackage{inconsolata}

\usepackage{graphicx}

\title{CoG: Controllable Graph Reasoning via Relational Blueprints and Failure-Aware Refinement over Knowledge Graphs}

\author{
    \textbf{Yuanxiang Liu}\textsuperscript{1},
    \textbf{Songze Li}\textsuperscript{1},
    \textbf{Xiaoke Guo}\textsuperscript{1},
    \textbf{Zhaoyan Gong}\textsuperscript{1},
    \textbf{Qifei Zhang}\textsuperscript{1}\thanks{~~Corresponding authors.}, \\ 
    \vspace{0.5em} 
    \textbf{Huajun Chen}\textsuperscript{1},
    \textbf{Wen Zhang}\textsuperscript{1}\footnotemark[1] \\ 
    \vspace{0.5em}
    \textsuperscript{1}Zhejiang University \\
    \vspace{0.5em}
    \texttt{\{liuyuanxiang, cstzhangqf, zhang.wen\}@zju.edu.cn}
}

\usepackage{enumitem}
\usepackage{tcolorbox}  
\tcbuselibrary{most}    
\usepackage{booktabs} 
\usepackage{multirow} 
\usepackage{amsmath}  
\usepackage{framed}  
\usepackage{listings}
\usepackage{xcolor}

\lstdefinelanguage{SPARQL}{
    morekeywords={PREFIX, SELECT, DISTINCT, WHERE, BIND, AS, UNION, FILTER, LANG, LANGMATCHES, ORDER, BY, ASC, LIMIT},
    morecomment=[l]{\#},
    morestring=[b]",
    sensitive=true
}

\newtcolorbox{sparqlbox}{
    colback=gray!5,       
    colframe=black,      
    boxrule=0.5pt,       
    sharp corners,       
    left=4pt,
    right=4pt,
    top=0pt,             
    bottom=2pt,
    before skip=6pt,
    after skip=6pt,
    fontupper=\small\ttfamily 
}

\begin{document}
\maketitle
\begin{abstract}
Large Language Models (LLMs) have demonstrated remarkable reasoning capabilities but often grapple with reliability challenges like hallucinations. While Knowledge Graphs (KGs) offer explicit grounding, existing paradigms of KG-augmented LLMs typically exhibit \textbf{cognitive rigidity}---applying homogeneous search strategies that render them vulnerable to instability under neighborhood noise and structural misalignment leading to reasoning stagnation. To address these challenges, we propose \textbf{CoG}, a training-free framework inspired by \textbf{Dual-Process Theory} that mimics the interplay between intuition and deliberation. First, functioning as the fast, intuitive process, the Relational Blueprint Guidance module leverages relational blueprints as interpretable soft structural constraints to rapidly stabilize the search direction against noise. Second, functioning as the prudent, analytical process, the Failure-Aware Refinement module intervenes upon encountering reasoning impasses. It triggers evidence-conditioned reflection and executes controlled backtracking to overcome reasoning stagnation. Experimental results on three benchmarks demonstrate that CoG significantly outperforms state-of-the-art approaches in both accuracy and efficiency.
\end{abstract}

\section{Introduction}
\label{sec:intro}

Large Language Models (LLMs)~\citep{io, Traininglanguagemodel, gpt4, llama, deepseek} have demonstrated strong generalization across natural language tasks~\citep{multitask, survey-llm, survey-reasoning, survey-prompting}. However, in knowledge-intensive and multi-hop reasoning scenarios, they face substantial reliability challenges, including hallucinations and inconsistent evidence chains~\citep{facts-check, Kd, llmcsc}. These issues are exacerbated by reliance on parametric knowledge, which is difficult to update and lacks verifiability. When external evidence is incomplete, models often fall back on language priors, producing answers that are fluent yet weakly grounded. Consequently, integrating knowledge graphs (KGs)---which offer structured, retrievable, and verifiable facts---has become a critical pathway to ground complex reasoning~\citep{roadmap-kg-llm}.

Among KG-augmented LLMs approaches, the \textbf{LLM-driven graph agent paradigm}~\citep{tog, pog, dog} has been widely adopted for its flexibility. Such methods typically execute a plan--retrieve--generate loop to iteratively extend evidence chains~\citep{tog, pog, rog}. Despite their utility, these approaches often exhibit instability in complex settings, with performance fluctuating heavily under neighborhood noise. We attribute this instability not merely to knowledge deficiency~\citep{survey-hallucination}, but to a lack of adaptive strategy regulation---a limitation we term \textbf{cognitive rigidity}. In practice, many existing systems~\citep{tog, pog} apply homogeneous search strategies regardless of task uncertainty, leading to reasoning trajectories that oscillate or deviate. Specifically, cognitive rigidity manifests in two reinforcing challenges (Figure~\ref{fig:case}):

\begin{figure*}[t]
  \centering
  \includegraphics[width=\textwidth]{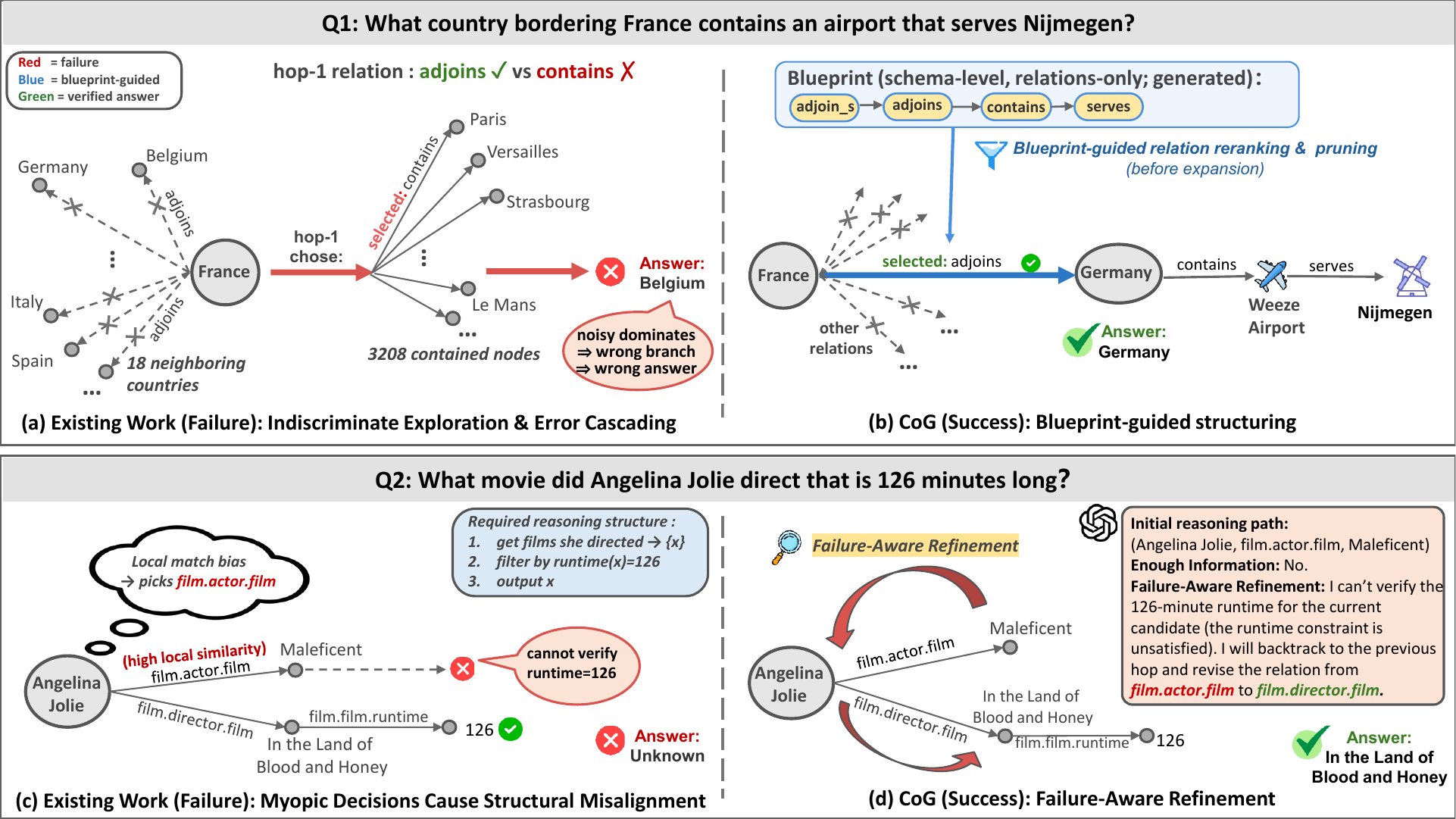}
  \vspace{-1.5em}
  \caption{Illustration of cognitive rigidity in existing works and how CoG addresses it:
  (I) \textbf{Error cascading} from indiscriminate exploration (top), mitigated by \textbf{Relational Blueprints} for blueprint-guided relation reranking and pruning (a vs.\ b); (II) \textbf{structural misalignment} from myopic decisions (bottom), corrected by \textbf{Failure-Aware Refinement} via backtracking and controlled fallback (c vs.\ d).}
  \label{fig:case}
  \vspace{-1em}
\end{figure*}

\smallskip
\noindent\textbf{(1) Error Cascading from Indiscriminate Exploration.}
Indiscriminate exploration fails to distinguish high-value signals from noise. A minor early selection error (e.g., selecting \textit{contains} instead of \textit{adjoins}) exposes the agent to substantially larger candidate sets. This noise dominates the reasoning branch, causing irreversible trajectory deviation (Figure~\ref{fig:case}(a)) that is difficult to recover from.

\smallskip
\noindent\textbf{(2) Structural Misalignment from Myopic Decisions.}
Relying heavily on local semantic matching often neglects global logical constraints. This myopia traps models in local optima: selecting relations that appear relevant but utilize the wrong structure (e.g., prioritizing \textit{actor} over \textit{director} in Figure~\ref{fig:case}(c)). Consequently, retrieved candidates fail to satisfy downstream constraints (e.g., runtime checks), forcing premature termination. This highlights that local semantic relevance does not guarantee cross-hop structural consistency.

To address these challenges, we propose \textbf{CoG}\footnote{\url{https://github.com/zjukg/CoG}.}, a novel training-free framework for \textbf{controllable reasoning on KGs}. Inspired by \textbf{Dual-Process Theory}~\citep{Thinking}, CoG emulates the cognitive interplay between intuitive pattern recognition and deliberate analysis: \textbf{First, Relational Blueprint Guidance (System~1):} Acting as the \textit{fast, intuitive} process driven by experience and structural patterns, this module is responsible for rapidly determining the search direction, particularly when structures are similar or information is sufficient. It leverages relational blueprints as heuristic priors to efficiently filter noise without heavy reasoning overhead (Figure~\ref{fig:case}(b)).
\textbf{Second, Failure-Aware Refinement (System~2):} Acting as the \textit{prudent, analytical} process, this module intervenes upon detecting failure signals, such as insufficient evidence or unverifiable constraints. It explicitly diagnoses anomalies (e.g., stagnation due to KG incompleteness~\citep{gog}) and executes corrective measures through controlled backtracking to overcome reasoning dead-ends (Figure~\ref{fig:case}(d)). Overall, CoG harmonizes intuitive guidance with analytical refinement to improve robustness while preserving verifiability. Our main contributions are:

\begin{itemize}

  \item We propose \textbf{CoG}, a training-free framework that synergizes blueprint-guided planning for stability and failure-aware refinement for robustness, effectively mitigating error cascading and structural mismatch.
  \item We introduce a relational blueprint mechanism, injecting interpretable soft structural priors into graph exploration to balance efficiency and controllability.
  \item We evaluate CoG on multiple KGQA datasets. Results on both \textbf{open-source and closed-source LLMs} validate that our framework outperforms state-of-the-art baselines and generalizes robustly across diverse backbones.
\end{itemize}

\section{Preliminaries}
\label{sec:preliminary}

\noindent\textbf{Knowledge Graph (KG).}\enspace
Let $\mathcal{G}=(\mathcal{E}, \mathcal{R})$ be a knowledge graph, where $\mathcal{E}$ and $\mathcal{R}$ denote entities and relations, respectively.
It consists of factual triplets $(e, r, e')$, where $e, e' \in \mathcal{E}$ and $r \in \mathcal{R}$.

\noindent\textbf{Relation \& Reasoning Paths.}\enspace
A \textbf{relation path} $z = (r_1, \dots, r_L)$ captures an abstract query pattern (e.g., \textit{born\_in} $\rightarrow$ \textit{capital\_of}) independent of concrete entities.
A \textbf{reasoning path} $p_z$ instantiates $z$ in $\mathcal{G}$ as $e_0 \xrightarrow{r_1} e_1 \dots \xrightarrow{r_L} e_L$, where each step $(e_{i-1}, r_i, e_i)$ is a valid triplet in $\mathcal{G}$.

\noindent\textbf{Problem Statement (KGQA).}\enspace
Given a question $Q$ and a KG $\mathcal{G}$, the goal is to identify the answer set $\mathcal{A} \subseteq \mathcal{E}$.
Following prior agent-based paradigms~\citep{tog, pog}, we adopt an iterative retrieve-and-reason approach: starting from topic entities $\mathcal{E}_q$ extracted from $Q$, the agent expands the subgraph step-by-step to construct a reasoning path pointing to the answer.

\begin{figure*}[t]
    \centering
    \includegraphics[width=1.0\textwidth]{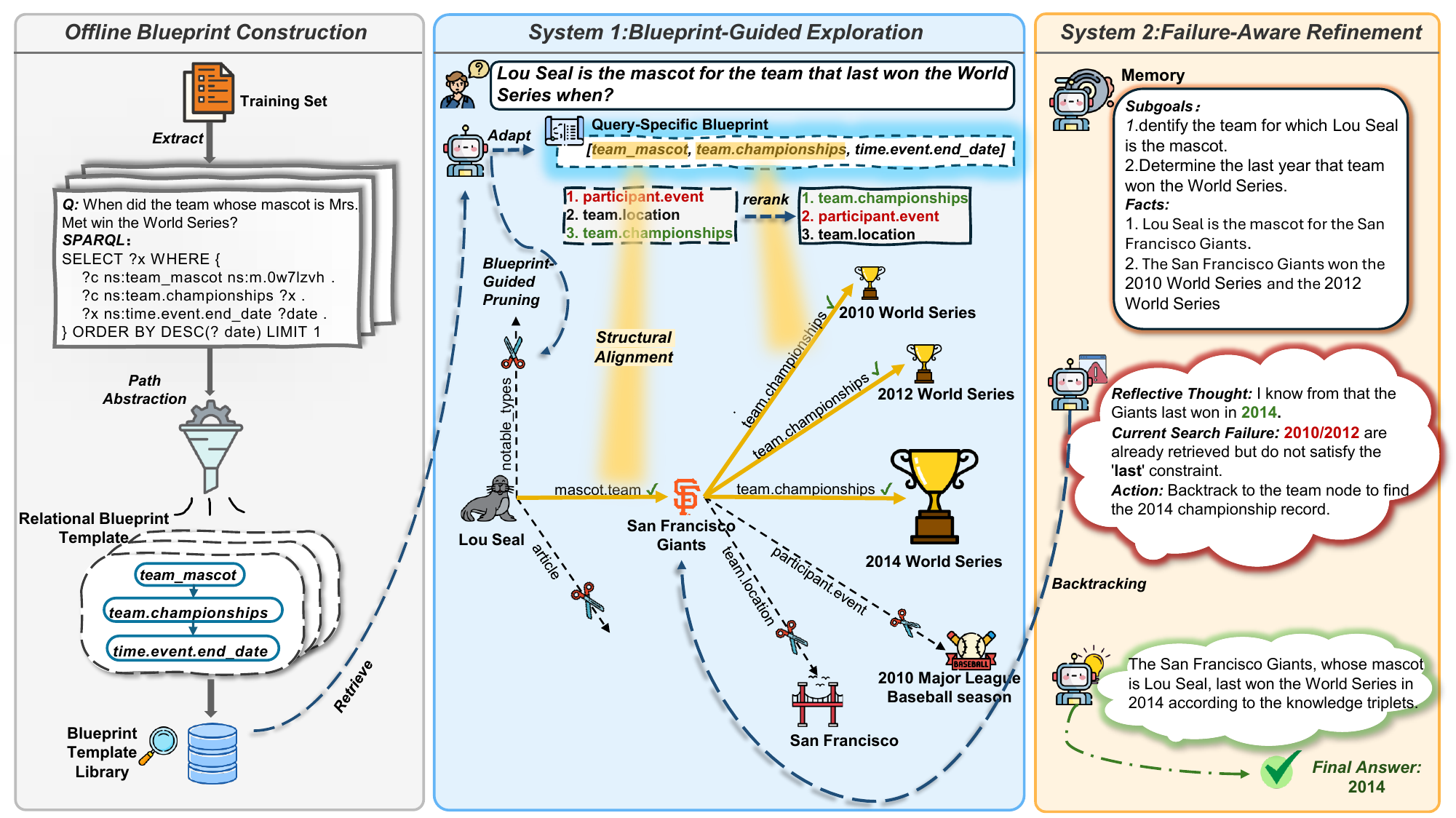}

    \caption{\textbf{Overview of the CoG framework.} CoG instantiates Dual-Process Theory as a cooperative reasoning loop. \textbf{Left:} Offline blueprint construction abstracts relational sequences from training paths into a searchable template library. \textbf{Middle (System 1):} Online blueprint-guided exploration adapts query-specific blueprints as soft structural constraints to rerank and prune candidate relations. \textbf{Right (System 2):} Failure-aware refinement monitors the reasoning process; upon failure signals (e.g., search stagnation), it performs evidence-conditioned reflection and targeted backtracking to recover a verifiable answer (e.g., enforcing the temporal constraint ``last won'').}

    \label{fig:overall_framework}
\end{figure*}

\section{Methodology}
\label{sec:method}

In this section, we introduce CoG, a training-free framework that couples blueprint-guided planning with failure-aware correction. As shown in Figure~\ref{fig:overall_framework}, CoG instantiates Dual-Process Theory with three components. Specifically, (i) Offline blueprint construction abstracts relational sequences from training paths into a searchable template library. (ii) System 1 (Online blueprint-guided exploration) retrieves and adapts a query-specific blueprint as a soft structural constraint for candidate-relation reranking and pruning. (iii) System 2 (Failure-aware refinement) detects failure signals (e.g., search stagnation) and triggers evidence-conditioned reflection with targeted backtracking to revise earlier decisions and recover a verifiable evidence chain. Overall, CoG promotes globally consistent exploration beyond purely local semantic matching.

\subsection{Offline Relational Blueprint Construction} 
\label{sec:offline}
In the offline stage, we distill structural priors from training data by abstracting entity-specific reasoning paths into reusable relation-only patterns. Crucially, this one-time preprocessing relies strictly on the standard training set, requiring zero external knowledge bases and zero LLM API calls. This yields a \emph{blueprint template library} that provides stable structural cues for online reasoning with negligible computational overhead.

\noindent\textbf{Extraction and Abstraction of Relation Paths.}
Given a training instance $(q,p)\in\mathcal{D}_{\text{train}}$, where $q$ is the question and $p$ is its reasoning-path representation (e.g., an executable SPARQL query), we aim to extract the underlying logical structure. Since $p$ is typically instantiated with concrete entities, it is not directly transferable. We therefore adopt a deterministic, rule-based de-instantiation strategy: we parse $p$ and apply pattern filters (e.g., regular expressions) to remove entity identifiers (e.g., Freebase IDs) and other non-structural elements, retaining only relation predicates. The remaining relations, ordered by occurrence, form a relational blueprint template:
\begin{equation}
\mathcal{S}(q)=\langle r_1,r_2,\ldots,r_L\rangle,\quad r_j\in\mathcal{R},
\end{equation}
where $\mathcal{R}$ is the set of KG relations, $r_j$ is the $j$-th predicate, and $L$ is the path length. This abstraction converts grounded facts into reusable structural priors that transfer across questions.

\noindent\textbf{Relational Blueprint Template Library Construction.}
Multiple training questions may share the same logical structure. To ensure a compact yet representative library, we deduplicate templates using their string representations as unique keys. For each unique $\mathcal{S}$, we aggregate its associated questions and select a \emph{semantic anchor} $q^{*}$ to facilitate online matching. We employ a simple yet effective heuristic by selecting the longest question as the anchor to preserve maximal contextual semantics:
\begin{equation}
q^{*}=\arg\max\nolimits_{q\in\{q' \mid \mathrm{map}(q')=\mathcal{S}\}} \mathrm{len}(q),
\end{equation}
where $\mathrm{map}(\cdot)$ is the abstraction mapping and $\mathrm{len}(\cdot)$ denotes text length. The final library is defined as $\mathcal{B}=\{(q_i,\mathcal{S}_i)\}_{i=1}^{N}$.

\noindent\textbf{Semantic Indexing.}
For efficient retrieval, we encode each anchor question $q_i$ with a pretrained sentence encoder (e.g., SentenceTransformer) to obtain $\mathbf{v}_{q_i}$, and build a vector index over $\{\mathbf{v}_{q_i}\}_{i=1}^{N}$ for nearest-neighbor search at test time. This offline procedure involves no task-specific training: we do not fine-tune the encoder or perform gradient updates, relying only on rule-based extraction and encoder forward passes.

\subsection{Online Blueprint-Guided KG Exploration}
The online stage incrementally constructs a traceable and verifiable evidence chain on the KG, while keeping reasoning stable and controllable under neighborhood noise and varying candidate scales. 
We instantiate CoG as a planning agent that performs dynamic navigation on the KG.
For a question $Q$, following prior work~\citep{tog, pog, kgagent} the agent follows an iterative loop: at step $t$, it identifies a candidate relation set $\mathcal{R}_{\text{cand}}$ reachable from the entity frontier $E_{t-1}$, executes a selection to expand the frontier to $E_t$, and verifies new evidence against constraints. Verified triples and intermediate conclusions are stored in a working memory $\mathcal{M}$, providing context for subsequent decisions.
On top of this loop, CoG retrieves and adapts a query-specific relational blueprint $S_{\text{BP}}$ and uses it as an interpretable soft structural constraint to guide candidate-relation reranking and pruning at each step, mitigating structural mismatch and error cascading from purely local semantic matching.

\subsubsection{Initialization and Structural Planning}

\noindent\textbf{Initialization and Task Decomposition.}
Given $Q$, we perform entity linking to identify topic entities and obtain the initial frontier $E_0$. A working memory $\mathcal{M}$ is maintained to store verified evidence triples, historical decisions, and constraint states, supporting conditional decision-making and failure correction. We then decompose $Q$ into an ordered subgoal sequence $\mathcal{O} = [o_1, \ldots, o_T]$, where each $o_t$ specifies the focus at step $t$ to mitigate context drift. While subgoals guide local actions, they do not define the global relational structure, which can lead to the structural misalignment and myopic decisions that CoG aims to resolve.

\noindent\textbf{Entity-masked blueprint retrieval and adaptation.}
To provide global guidance, we generate a query-specific blueprint $S_{\text{BP}}$. We first mask topic entities in $Q$ to obtain $Q_m=\mathrm{Mask}(Q,E_0)$, shifting retrieval toward compositional structures and predicate patterns.
We encode $Q_m$ with a pretrained sentence encoder $f(\cdot)$ and retrieve Top-$K$ nearest exemplars from an offline semantic index, where each exemplar is an \emph{anchor-question--relational-blueprint-template} pair. We adopt a hybrid \emph{copy--adapt} strategy: if the top similarity exceeds $\tau_{\text{copy}}$, we copy the retrieved Top-1 template; otherwise, we feed the most similar exemplars (e.g., Top-2) together with $Q$ into an LLM to generate or lightly rewrite a blueprint under exemplar structural constraints. Crucially, this adaptation mechanism ensures that CoG is not confined to the training distribution. By treating retrieved blueprints as flexible structural exemplars rather than rigid rules, the LLM can synthesize novel reasoning plans for unseen query topologies, effectively bridging the gap between historical priors and new scenarios. Specific implementation details are provided in Appendix~\ref{sec:Sensitivity_of_Copy_Threshold}. 
The resulting blueprint is
\begin{equation}
S_{\text{BP}}=\langle r^{\text{BP}}_1,\ldots,r^{\text{BP}}_L\rangle,
\end{equation}
where $r^{\text{BP}}_j$ is the $j$-th relation slot and $L$ is the blueprint length. Importantly, $S_{\text{BP}}$ is not a hard executable plan but an interpretable soft structural prior: it sketches the approximate relation types to follow and continuously constrains candidate selection. $L$ provides a depth prior (not a hard limit); the actual number of steps is determined by constraint satisfaction and termination criteria.

\subsubsection{Blueprint-Guided KG Exploration}

Under subgoals and the blueprint, CoG performs stepwise exploration by expanding the frontier, verifying evidence, and updating memory under the blueprint soft constraint.

\noindent\textbf{Candidate relation collection and blueprint-guided reranking.}
From the current frontier $E_{t-1}$, we collect all reachable relations to form the candidate set $\mathcal{R}_{\text{cand}}$.
To control scale and suppress noise propagation, we apply lightweight rule-based filtering (e.g., removing obviously uninformative generic relations).
In noisy neighborhoods, $\mathcal{R}_{\text{cand}}$ may still contain many distractors, and relying only on local semantic matching can cause early mis-selections and error cascading.
To stabilize decisions, CoG injects the blueprint $S_{\text{BP}}$ into the candidate-relation layer and reranks candidates by three complementary signals.
We first define a slot-alignment index
\begin{equation}
\pi(t)=\arg\max\nolimits_{j\in[1,L]} \mathrm{sim}\!\big(h(o_t),h(r^{\text{BP}}_j)\big),
\end{equation}
where $h(\cdot)$ is a text encoder and $\mathrm{sim}(\cdot,\cdot)$ is cosine similarity.
To prevent structural drift caused by out-of-order slot alignment in multi-hop reasoning, we enforce a monotone alignment constraint in implementation:
\begin{itemize}[leftmargin=*, nosep]
    \item \textbf{Initialization:} $\pi(0)=1$.
    \item \textbf{Monotone update:} restrict $j\in[\pi(t-1),L]$ so $\pi(t)$ is non-decreasing.
    \item \textbf{Clamping:} when steps exceed $L$, clamp $\pi(t)=L$.
\end{itemize}
Based on $\pi(t)$, we define three complementary scoring signals:
\begin{subequations}
\fontsize{10pt}{11pt}\selectfont
\begin{align}
\phi_{\mathrm{loc}}(o_t,r) &= \mathrm{sim}\!\big(h(o_t),h(r)\big), \\
\phi_{\mathrm{step}}(r,r^{\text{BP}}_{\pi(t)}) &= \mathrm{sim}\!\big(h(r),h(r^{\text{BP}}_{\pi(t)})\big), \\
\phi_{\mathrm{glob}}(S_{\text{BP}},r) &= \max\nolimits_{j\in[1,L]} \mathrm{sim}\!\big(h(r),h(r^{\text{BP}}_j)\big),
\end{align}
\end{subequations}
where $\phi_{\mathrm{loc}}$ (\emph{local relevance}) captures subgoal--relation relevance,
$\phi_{\mathrm{step}}$ (\emph{step-wise alignment}) measures alignment to the current blueprint slot,
and $\phi_{\mathrm{glob}}$ (\emph{global compatibility}) evaluates compatibility with the overall blueprint, helping mitigate structural drift over long-horizon reasoning. For each $r\in\mathcal{R}_{\text{cand}}$, we compute a fused score
\begin{equation}
\begin{split}
\mathrm{Score}(r) &= \lambda_{\mathrm{loc}}\phi_{\mathrm{loc}}(o_t,r) \\ &\quad + \lambda_{\mathrm{step}}\phi_{\mathrm{step}}(r,r^{\text{BP}}_{\pi(t)}) \\ &\quad + \lambda_{\mathrm{glob}}\phi_{\mathrm{glob}}(S_{\text{BP}},r),
\end{split}
\end{equation}
where $\lambda_{\mathrm{loc}}, \lambda_{\mathrm{step}}, \lambda_{\mathrm{glob}} \ge 0$ and
$\lambda_{\mathrm{loc}}+\lambda_{\mathrm{step}}+\lambda_{\mathrm{glob}}=1$. we set weights $\lambda_{\mathrm{loc}}{=}0.6$, $\lambda_{\mathrm{step}}{=}0.25$, and $\lambda_{\mathrm{glob}}{=}0.15$ in all experiments; additional sensitivity details are deferred to Appendix~\ref{sec:sensitivity_reranking}. We then rerank $\mathcal{R}_{\text{cand}}$ by $\mathrm{Score}(\cdot)$ and retain the top-scoring relations to form a compact, structure-aligned shortlist $\tilde{\mathcal{R}}_{\text{cand}}$.

\noindent\textbf{Blueprint-guarded pruning.}
We further refine $\tilde{\mathcal{R}}_{\text{cand}}$ via an LLM conditioned on $(Q,o_t,\mathcal{M})$ and the shortlist. To reduce the risk of missing structurally correct relations, we enforce a Structure-Consistency Safeguard: the final candidate set is the \textbf{union} of the LLM-selected relations and the top candidate by step-wise alignment $\phi_{\mathrm{step}}$. This dual-source selection balances semantic nuances and structural consistency, mitigating single-view bias.

\noindent\textbf{State update and answer generation.}
After selecting $r_t$, CoG expands the KG to obtain the next frontier $E_t$, filters reached entities using subgoal constraints and the working memory $\mathcal{M}$ (via an LLM), and writes the chosen relation, key evidence triples, and verified intermediate conclusions into $\mathcal{M}$ for traceability and termination. CoG then performs an LLM-based sufficiency check conditioned on the current subgoal state $o_t$ and verified evidence in $\mathcal{M}$. If sufficient, the LLM synthesizes the verified traces and subgoal states to produce the final answer. If insufficient, the failure stems from either missing evidence (addressable by further expansion) or an early wrong decision that derailed the trajectory; in the latter case, CoG invokes Failure-Aware Refinement to backtrack and revise suspicious transitions, recovering a verifiable evidence chain while maintaining controllability.

\subsection{Failure-Aware Refinement}
While blueprint guidance provides strong structural priors, KG incompleteness and noise can still impede exploration. To counteract this, CoG implements Failure-Aware Refinement. Upon detecting failure signals (e.g., stagnation or insufficient evidence), CoG switches from forward exploration to a controlled correction mode. This process prioritizes minimal interventions on high-risk transitions, ensuring the recovery of verifiability without discarding valid sub-paths.

\noindent\textbf{Diagnosis and Targeted Re-routing.}
Upon triggering a failure signal, CoG invokes an LLM for evidence-conditioned reflection. Guided by the working memory $\mathcal{M}$, the the LLM reviews the current trajectory $\mathcal{T}=[e_0,r_1,e_1,\ldots]$ alongside compact summaries of pruned branches to pinpoint the decision point $t_{\text{err}}$ responsible for the deviation (e.g., a step biased by local semantics).
Subsequently, the agent executes targeted backtracking to re-route the search: it reverts the frontier to the state preceding $t_{\text{err}}$, recalls structurally relevant candidates that were prematurely pruned, and resumes expansion. This re-routing mechanism rectifies myopic errors while enforcing global constraints, allowing CoG to reconstruct a verifiable evidence chain.

\noindent\textbf{Grounded Inference.}
In extreme cases, missing edges can render verifiable evidence unreachable, such that even re-routing fails to restore a checkable evidence chain. As a fallback, CoG aggregates verified path segments and unmet constraints into a concise summary, prompting the LLM to synthesize a final answer conditioned strictly on this valid context. Unlike free-form generation, this inference is explicitly grounded in the verified semantic space, effectively mitigating parametric hallucination while preventing premature termination. For a deeper empirical understanding, a comprehensive quantitative and qualitative analysis of this refinement module---demonstrating how it successfully diagnoses structural misalignment and recovers from search impasses---is detailed in Appendix~\ref{app:refinement}.

\section{Experiments}
\label{sec:experiments}

\subsection{Experiment Setup}
\label{sec:exp_setup}

\noindent\textbf{Datasets \& Evaluation Metrics.} 
To verify CoG's effectiveness in complex reasoning over knowledge graphs, \textbf{aligning with the standard evaluation protocols of prior state-of-the-art methods}~\citep{tog, pog}, we evaluate on three representative multi-hop KGQA benchmarks: \textbf{CWQ}~\citep{cwq}, \textbf{WebQSP}~\citep{webqsp}, and \textbf{GrailQA}~\citep{grailqa}. 
These datasets are grounded on Freebase~\citep{freebase}, which contains approximately 88 million entities, 20K relations, and 126 million triples, serving as one of the most comprehensive knowledge bases for standardized KGQA evaluation. Further details of these datasets are provided in Appendix~\ref{sec:datasets}. 
Following previous studies~\citep{tog, pog, structgpt}, we use Exact Match accuracy (\textbf{Hits@1}) as the primary metric to evaluate whether the predicted answer exactly matches the ground-truth. Furthermore, to comprehensively capture the model's ability to retrieve complete answer sets for complex queries where multiple valid entities exist, we incorporate the \textbf{F1-score} to measure recall and overall coverage.

\noindent\textbf{Compared Methods.}
We compare CoG with strong existing LLM-based baselines from two groups: LLM-only methods and KG-augmented LLM methods, covering both fine-tuned and prompting methods. Detailed descriptions of the compared methods are deferred to Appendix~\ref{sec:baseline_descriptions}.

\noindent\textbf{Implementations.} 
We construct dataset-specific blueprint libraries by parsing gold SPARQL queries from the training splits of WebQSP, GrailQA, and CWQ. Semantic anchors are encoded into dense vectors using a pre-trained Sentence Transformer~\citep{distilbert} for retrieval. Detailed implementation settings and template library statistics are deferred to the Appendix~\ref{sec:Blueprint Library Statistics}.

\subsection{Main Results}

\begin{table*}[t]
\centering
\small
\setlength{\tabcolsep}{12pt} 
\renewcommand{\arraystretch}{1.05} 

\begin{tabular}{lcccc}
\toprule
\multirow{2}{*}{\textbf{Method}} & \multicolumn{2}{c}{\textbf{CWQ}} & \multicolumn{2}{c}{\textbf{WebQSP}} \\
\cmidrule(lr){2-3} \cmidrule(lr){4-5}
& \textbf{Hits@1} & \textbf{F1} & \textbf{Hits@1} & \textbf{F1} \\
\midrule

\multicolumn{5}{c}{\textit{LLM-Only (GPT-3.5)}} \\
\midrule
IO Prompt~\citep{io} & 37.6 & -- & 63.3 & -- \\
CoT~\citep{cot}      & 38.8 & -- & 62.2 & -- \\
SC~\citep{sc}        & 45.4 & -- & 61.1 & -- \\
\midrule

\multicolumn{5}{c}{\textit{Fine-Tuned KG-Augmented LLM}} \\
\midrule
RE-KBQA~\citep{rekbqa}   & 50.3 & -- & 74.6 & -- \\
UniKGQA~\citep{unikgqa}  & 51.2 & 48.0 & 77.2 & 72.2 \\
RoG~\citep{rog}          & 62.6 & 56.2 & 85.7 & 70.8 \\
DECAF~\citep{decaf}      & 70.4 & -- & 82.1 & -- \\
KG-Agent~\citep{kgagent} & 72.2 & -- & 83.3 & -- \\
\midrule

\multicolumn{5}{c}{\textit{Prompting KG-Augmented LLM w/ Qwen2.5-7B}} \\
\midrule
ToG~\citep{tog}      & 42.5 & 28.7 & 56.0 & 37.3 \\
PoG~\citep{pog}      & 46.0 & 31.4 & 58.5 & 40.4 \\
\textbf{CoG (Ours)}  & \textbf{54.0} & \textbf{48.5} & \textbf{74.5} & \textbf{63.9} \\
\midrule

\multicolumn{5}{c}{\textit{Prompting KG-Augmented LLM w/ GPT-3.5 \& Others}} \\
\midrule
KD-CoT~\citep{Kd} (GPT-3.5 Turbo)        & 50.5 & -- & 73.7 & 50.2 \\
KB-BINDER~\citep{kbbinder} (Codex)       & -- & -- & 74.4 & -- \\
StructGPT~\citep{structgpt} (GPT-3)      & 54.3 & 49.6 & 72.6 & 63.7 \\
ToG~\citep{tog} (GPT-3.5 Turbo)          & 57.1 & 41.9 & 76.2 & 50.9 \\
PoG~\citep{pog} (GPT-3.5 Turbo)          & 63.2 & 43.7 & 82.0 & 58.1 \\
\textbf{CoG (Ours) (GPT-3.5 Turbo)}      & \textbf{66.9} & \textbf{59.9} & \textbf{86.8} & \textbf{74.3} \\
\midrule

\multicolumn{5}{c}{\textit{Prompting KG-Augmented LLM w/ GPT-4}} \\
\midrule
ToG~\citep{tog}      & 67.6 & 47.6 & 82.6 & 58.9 \\
PoG~\citep{pog}      & 75.0 & 42.1 & 87.3 & 59.8 \\
\textbf{CoG (Ours)}  & \textbf{77.8} & \textbf{69.2} & \textbf{89.7} & \textbf{75.5} \\
\bottomrule
\end{tabular}

\vspace{-10pt}
\caption{Performance comparison of different methods on CWQ and WebQSP. All baseline results are cited directly from original papers to ensure fair comparison on identical KG snapshots, mitigating variance from local reproductions.}
\label{tab:main_results}
\vspace{-10pt} 
\end{table*}

\begin{table}[t]
\centering
\setlength{\tabcolsep}{1.5pt}

\resizebox{\columnwidth}{!}{
    \begin{tabular}{lcccc}
    \toprule
    \multirow{2}{*}{\textbf{Method}} & \multicolumn{4}{c}{\textbf{GrailQA}} \\
    \cmidrule(lr){2-5}
    & \textbf{Overall} & \textbf{I.I.D.} & \textbf{Compositional} & \textbf{Zero-shot} \\
    \midrule

    \multicolumn{5}{c}{\textit{LLM-Only}} \\
    \midrule
    IO Prompt~\citep{io} & 29.4 & -- & -- & -- \\
    CoT~\citep{cot}            & 28.1 & -- & -- & -- \\
    SC~\citep{sc}             & 29.6 & -- & -- & -- \\
    \midrule

    \multicolumn{5}{c}{\textit{Fine-Tuned KG-Augmented LLM}} \\
    \midrule
    RnG-KBQA~\citep{rngkbqa}  & 68.8 & 86.2 & 63.8 & 63.0 \\
    TIARA~\citep{tiara}  & 73.0 & 87.8 & 69.2 & 68.0 \\
    FC-KBQA~\citep{fckbqa} & 73.2 & 88.5 & 70.0 & 67.6 \\
    Pangu~\citep{pangu}    & 75.4 & 84.4 & 74.6 & 71.6 \\
    FlexKBQA~\citep{flexkbqa} & 62.8 & 71.3 & 59.1 & 60.6 \\
    GAIN~\citep{gain} & 76.3 & 88.5 & 73.7 & 71.8 \\
    KG-Agent~\citep{kgagent}    & 86.1 & 92.0 & 80.0 & 86.3 \\
    \midrule

    \multicolumn{5}{c}{\textit{Prompting KG-Augmented LLM w/ Qwen2.5-7B}} \\
    \midrule
    ToG~\citep{tog}       & 62.6 & 60.8 & 47.0 & 68.9 \\
    PoG~\citep{pog}       & 68.9 & 67.9 & 51.5 & 75.4 \\
    \textbf{CoG (Ours)}  & \textbf{72.0} & \textbf{72.5} & \textbf{57.6} & \textbf{76.9} \\
    \midrule

    \multicolumn{5}{c}{\textit{Prompting KG-Augmented LLM w/ GPT-3.5 or others}} \\
    \midrule
    KB-BINDER~\citep{kbbinder} & 53.2 & 72.5 & 51.8 & 45.0 \\
    ToG~\citep{tog}              & 68.7 & 70.1 & 56.1 & 72.7 \\
    PoG~\citep{pog}              & 76.5 & 76.3 & 62.1 & 81.7 \\
    \textbf{CoG (Ours)}        & \textbf{79.2} & \textbf{80.4} & \textbf{65.2} & \textbf{83.6} \\
    \midrule

    \multicolumn{5}{c}{\textit{Prompting KG-Augmented LLM w/ GPT-4}} \\
    \midrule
    ToG~\citep{tog}       & 81.4 & 79.4 & 67.3 & 86.5 \\
    PoG~\citep{pog}       & 84.7 & 87.9 & 69.7 & 88.6 \\
    \textbf{CoG (Ours)}  & \textbf{86.4} & \textbf{88.3} & \textbf{76.3} & \textbf{89.1} \\
    \bottomrule
    \end{tabular}
}
\vspace{-10pt}
\caption{Performance comparison of different methods on GrailQA.}
\label{tab:grailqa}
\vspace{-10pt} 
\end{table}

Table~\ref{tab:main_results} and Table~\ref{tab:grailqa} present the comparisons between CoG and representative baselines on WebQSP, CWQ, and GrailQA. Overall, CoG delivers consistently strong performance across all evaluation settings. 

First, compared with prompting-based KG-augmented LLM baselines, CoG demonstrates clear superiority. Under both GPT-3.5 and GPT-4 backbones, CoG consistently outperforms the strongest baseline, PoG. While PoG utilizes adaptive planning, its search remains largely driven by local exploration signals, leaving it vulnerable to structural ambiguity in complex neighborhoods. CoG mitigates this limitation through the synergy of relational blueprint guidance and failure-aware refinement, enforcing global structural alignment and controlled error recovery to achieve higher accuracy. Furthermore, to evaluate the retrieval of complete answer sets, we introduce the F1-score metric. As shown in Table~\ref{tab:main_results}, CoG yields substantial F1 improvements—such as a \textbf{+27.1 absolute margin} over PoG on CWQ using GPT-4. This confirms that CoG effectively retrieves comprehensive answer sets rather than isolated entities, mitigating premature termination and ensuring superior coverage.

Second, although CoG is a training-free prompting framework, it remains highly competitive against fine-tuned KG-augmented LLM baselines. On CWQ and WebQSP, CoG powered by GPT-4 surpasses all included fine-tuned baselines; remarkably, on GrailQA, it maintains this superiority even when restricted to the less capable GPT-3.5 backbone. A critical concern regarding blueprint-guided exploration is its behavior in out-of-distribution settings involving unseen structures. However, evaluations on the GrailQA Zero-shot split (Table~\ref{tab:grailqa}) using GPT-3.5 demonstrate that CoG achieves an accuracy of 83.6\%, substantially outperforming both the exploration-based ToG at 72.7\% and the adaptive PoG at 81.7\%. This empirically confirms that our approach leverages training data as abstract logical priors to aid generalization, rather than merely memorizing paths, thereby offering superior robustness even with less capable backbone models. Moreover, these findings strongly support the effectiveness of combining structural guidance with failure awareness for complex reasoning, and further suggest that explicit structural priors coupled with dynamic self-correction can generalize more robustly to unseen query structures and compositional patterns than relying solely on implicit parameter updates.

Furthermore, compared to LLM-only baselines, CoG achieves substantial gains by integrating structured evidence from external knowledge graphs. This underscores the necessity of KG grounding for complex queries where parametric knowledge alone falls short. To evaluate generalization across model scales, we deploy CoG on a smaller backbone, Qwen2.5-7B~\citep{qwen}. Even with significantly reduced parameter capacity, CoG maintains its superiority over baseline methods. This confirms that the observed improvements stem primarily from our methodological innovations rather than mere model scaling. Finally, to qualitatively illustrate how CoG navigates complex reasoning paths and dynamically corrects errors, we provide detailed case studies in Appendix~\ref{sec:case}.

\subsection{Ablation Study}
\label{sec:ablation}

\begin{table}[t]
\centering
\setlength{\tabcolsep}{2pt}

\resizebox{\columnwidth}{!}{
    \begin{tabular}{lccc}
    \toprule
    \textbf{Method} & \textbf{CWQ} & \textbf{WebQSP} & \textbf{GrailQA} \\
    \midrule
    \textbf{CoG (Ours)} & \textbf{66.9} & \textbf{86.8} & \textbf{79.2} \\
    \midrule
    \multicolumn{4}{l}{\textit{Component Removal}} \\
    w/o Blueprint Adaptation & 62.4 & 83.5 & 77.5 \\
    w/o Blueprint-guided Reranking & 63.5 & 84.0 & 76.8 \\
    w/o Failure-Aware Refinement & 58.5 & 79.9 & 75.3 \\
    \textbf{w/o Blueprint Guidance (System 2 Only)} & 61.5 & 82.2 & 76.4 \\
    \midrule
    \multicolumn{4}{l}{\textit{Reranking Variants}} \\
    ~~Local relevance only & 64.6 & 84.4 & 76.2 \\
    ~~w/o Global compatibility & 65.8 & 85.4 & 77.2 \\
    ~~w/o Step-wise alignment & 65.9 & 85.7 & 77.6 \\
    \bottomrule
    \end{tabular}
}
\vspace{-10pt}
\caption{Ablation study of core components and blueprint-guided reranking signals.}
\label{tab:ablation}
\vspace{-10pt} 
\end{table}

To evaluate the effectiveness of individual components and our dual-process design, we conduct extensive ablation studies on CWQ, WebQSP, and GrailQA, as shown in Table~\ref{tab:ablation}. Specifically, \emph{w/o Blueprint Adaptation} refers to using retrieved relations directly from the database without allowing the LLM to modify them. \emph{w/o Blueprint-guided Reranking} removes the step where the blueprint serves as an interpretable structural prior for candidate relation selection. \emph{w/o Failure-Aware Refinement} disables the failure recovery mechanism, preventing the agent from correcting its reasoning when search stalls. Furthermore, to explicitly decouple the contributions of our framework, we introduce a \emph{w/o Blueprint Guidance (System 2 Only)} baseline, where the agent explores purely via failure-aware refinement, completely stripped of initial structural priors.

Empirical results reveal several key insights. First, removing \emph{Failure-Aware Refinement} causes the most significant performance drop, underscoring its critical role in resolving search impasses. Second, removing overall blueprint guidance (\emph{System 2 Only}) leads to a substantial decrease (e.g., a \textbf{5.4\% absolute drop} on CWQ), decisively quantifying the value of System 1 in providing strategic direction and preventing search myopia. Crucially, however, even this System 2 Only baseline (61.5\% on CWQ) still significantly outperforms pure exploration agents like ToG (57.1\%). This confirms that while System 1 blueprints are essential for peak performance, System 2 independently establishes a highly resilient reasoning foundation. Moreover, removing either \emph{Blueprint Adaptation} or \emph{Blueprint-guided Reranking} consistently degrades performance, indicating that CoG benefits from both adapting templates and leveraging them for ranking. Notably, while the blueprint offers implicit global context even without reranking, its efficacy is substantially diminished when decoupled from candidate ranking. Regarding reranking variants, relying solely on \emph{local relevance} yields suboptimal gains. Incorporating \emph{Step-wise Alignment} mitigates myopic decision-making, while \emph{Global Compatibility} ensures long-term consistency. Peak performance is achieved by integrating all three signals, confirming the synergy of our fused ranking design.

\subsection{Efficiency Analysis}
\label{sec:Efficiency Analysis}

\begin{table}[t]
\centering
\setlength{\tabcolsep}{2.5pt}

\resizebox{\columnwidth}{!}{
    \begin{tabular}{lcccccc}
    \toprule
    \textbf{Data} & \textbf{Method} & \textbf{Calls} & \textbf{Input} & \textbf{Output} & \textbf{Total} & \textbf{H@1} \\
    \midrule
    
    \multirow{3}{*}{\textbf{CWQ}} 
    & ToG & 22.6 & 8,182.9 & 1,486.4 & 9,669.4 & 57.1 \\
    & PoG & 13.3 & 7,803.0 & \textbf{353.2} & 8,156.2 & 63.2 \\
    & \textbf{CoG} & \textbf{11.7} & \textbf{6,589.0} & 486.8 & \textbf{7,075.8} & \textbf{66.9} \\
    \midrule
    
    \multirow{3}{*}{\textbf{WebQSP}} 
    & ToG & 15.9 & 6,031.2 & 987.7 & 7,018.9 & 76.2 \\
    & PoG & 9.0 & 5,234.8 & 282.9 & 5,517.7 & 82.0 \\
    & \textbf{CoG} & \textbf{8.3} & \textbf{4,693.6} & \textbf{206.0} & \textbf{4,899.6} & \textbf{86.8} \\
    \midrule
    
    \multirow{3}{*}{\textbf{GrailQA}} 
    & ToG & 11.1 & 4,066.0 & 774.6 & 4,840.6 & 68.7 \\
    & PoG & 6.5 & 3,372.8 & 202.8 & 3,575.6 & 76.5 \\
    & \textbf{CoG} & \textbf{5.5} & \textbf{3,122.0} & \textbf{166.1} & \textbf{3,288.1} & \textbf{79.2} \\
    \bottomrule
    \end{tabular}
}
\vspace{-10pt}
\caption{Efficiency comparison. Metrics include average LLM calls and token usage per query.}
\label{tab:efficiency}
\vspace{-10pt} 
\end{table}

We evaluate the efficiency of different methods in terms of LLM calls and token usage, as shown in Table~\ref{tab:efficiency}. CoG achieves a superior accuracy-cost trade-off across all datasets. Unlike traditional expansion-based methods, which rely on exhaustive searches over neighboring nodes, CoG uses the relational blueprint to constrain the search space. This reduces irrelevant reasoning paths by pruning them in advance, eliminating the need for exhaustive exploration and avoiding redundant intermediate queries. As a result, CoG significantly reduces computational overhead, setting a new performance benchmark while minimizing costs, making it highly cost-effective and practical for real-world deployment.

\section{Related Work}
\label{sec:Related Work}

\noindent\textbf{LLM Reasoning.} 
LLMs have transitioned from heuristic prompting to structured reasoning frameworks~\citep{io, cot, ltm, survey-llm, survey-reasoning, survey-prompting}. Beyond Chain-of-Thought (CoT)~\citep{cot}, recent works utilize non-linear topologies~\citep{tot, got, mot, sot} to navigate complex spaces~\citep{tot, got}. However, purely parametric reasoning remains vulnerable to hallucinations and logical inconsistencies~\citep{survey-hallucination, facts-check}. While ensemble strategies like majority voting~\citep{sc} bolster stability, they cannot rectify knowledge staleness or decision opacity~\citep{survey-reasoning, roadmap-kg-llm}.

\noindent\textbf{KG-Augmented LLM.} 
To overcome parametric memory limits, recent efforts ground LLMs in diverse external evidence, from tabular data~\citep{astra} to dynamic temporal KGs~\citep{rtqa, tempr1}. Within this context, standard KGs offer verifiable grounding and structural control~\citep{survey-hallucination, roadmap-kg-llm, chen2024large} via implicit integration~\citep{unikgqa, rog} or explicit retrieval~\citep{kbbinder, structgpt, eog}. Yet, implicit methods struggle with schema evolution, and explicit retrieval often decouples search from reasoning, requiring multi-stage noise filtering~\citep{structgpt, kbbinder, huang2025general}. Agent-based paradigms~\citep{tog, pog} enable interactive navigation but risk irreversible semantic drift. Specifically, although Plan-on-Graph (PoG)~\citep{pog} uses adaptive planning, its purely ``on-the-fly'' local exploration lacks global foresight, making it highly susceptible to structural ambiguity. Conversely, while recent studies like GCR~\citep{gcr} employ rigid KG-tries for stronger regulation and robust feature interactions~\citep{wang2024gate}, these hard branching constraints are brittle in incomplete real-world KGs; a single missing edge causes branch collapse.

In contrast, CoG's closed-loop dual-process paradigm effectively mitigates PoG's local myopia and GCR's brittleness. Unlike PoG's purely online planning, CoG extracts structural knowledge offline into a global reference library. This ``compass'' conceptualizes relational blueprints as \emph{soft structural priors} (System 1)~\citep{wang2024embedding}, seamlessly blending textual intension with structural extension. Dynamic branching and error-recovery are instead intentionally offloaded to Failure-Aware Refinement (System 2). This design strikes a superior balance: avoiding parameter fine-tuning, preventing aimless wandering and computational bottlenecks of zero-shot exploration~\citep{zhou2024research}, and ensuring robust recovery from local ambiguities and missing edges, while preserving adaptive flexibility.

\section{Conclusion}
\label{sec:Conclusion}
In this paper, we present CoG, a training-free, dual-process framework for controllable reasoning over knowledge graphs. By seamlessly integrating relational blueprint-guided exploration (System 1) with failure-aware refinement (System 2), CoG fundamentally mitigates critical challenges such as search myopia, error cascading, and structural misalignment in complex multi-hop reasoning. Extensive experiments across WebQSP, CWQ, and GrailQA demonstrate that CoG consistently achieves state-of-the-art performance. Most notably, it excels in zero-shot generalization and the retrieval of comprehensive answer sets without relying on costly parameter fine-tuning. Ultimately, CoG establishes a highly robust, interpretable, and scalable paradigm for knowledge graph question answering.

\section*{Limitations}

Despite the effectiveness of CoG in enhancing multi-hop reasoning, several limitations present promising avenues for future work. First, its absolute performance is inherently bounded by the completeness of the underlying knowledge graph. While failure-aware refinement mitigates missing edges locally, it cannot fully compensate for entirely absent reasoning paths. Second, the effectiveness of structural guidance relies on the coverage of the pre-constructed blueprint library; in highly niche domains with scarce training queries, retrieving adaptable templates may prove challenging. Third, while CoG achieves a highly favorable accuracy-cost trade-off, resolving complex cascading failures through iterative backtracking naturally introduces additional computational latency compared to single-pass inference. Finally, while our linear blueprints serve as highly robust soft priors against KG incompleteness, exploring more expressive structural formats---such as hybridizing our dynamic refinement mechanism with tree-structured constraints (e.g., KG-tries~\citep{gcr})---represents an intriguing direction for exceptionally compositional queries. Additionally, transitioning from static offline blueprint generation to dynamic online evolution could further enhance the framework's adaptability.

\section*{Acknowledgements}

This work is founded by National Natural Science Foundation of China (NSFC62306276/\allowbreak NSFCU23B2055), New Generation Artificial Intelligence-\allowbreak National Science and Technology Major Project 2030 (2025ZD0122800), Yongjiang Talent Introduction Programme (2022A-238-G), and Fundamental Research Funds for the Central Universities (226-2023-00138). This work was supported by Ant Group.

\bibliography{custom}

\appendix

\section{Prompt Templates for LLM Agents}
\label{sec:PromptTemplates}

In this section, we provide the prompt templates used in CoG for LLM agents. To ensure proper parsing of the LLM output, we require the LLM to provide answers using specific data structures, such as lists and JSON. This structured output facilitates easier processing and integration with the system. Additionally, we explicitly instruct the LLM not to include any irrelevant information in its responses. Some of the prompt templates are based on the design of PoG (Plan-on-Graph)~\citep{pog} prompts, and have been adjusted and optimized to better suit the reasoning process in CoG.

\subsection{Task Decomposition}
\label{sec:task_decomp_prompt}

\begin{tcolorbox}[
    colback=white, 
    colframe=black, 
    boxrule=0.5pt, 
    sharp corners,
    left=4pt,   
    right=4pt, 
    top=0pt,    
    bottom=2pt, 
    before skip=6pt,
    after skip=6pt
]
\small
\noindent Please break down the process of answering the question into as few subgoals as possible based on semantic analysis. \\

\noindent \textit{In-Context Few-Shot} \\

\noindent Now you need to directly output subgoals of the following question in list format without other information or notes. \\
\noindent Q: \{\}
\end{tcolorbox}

\subsection{KG Exploration}
\label{sec:kg_exploration_prompts}

\subsubsection{Relation Pruning}
\label{sec:relation_pruning_prompt}

\begin{tcolorbox}[
    colback=white, 
    colframe=black, 
    boxrule=0.5pt, 
    sharp corners,
    left=4pt,   
    right=4pt, 
    top=0pt,    
    bottom=2pt, 
    before skip=6pt,
    after skip=6pt
]
\small
\noindent Please provide as few highly relevant relations as possible to the question and its subgoals from the following relations (separated by semicolons). \\

\noindent \textit{In-Context Few-Shot} \\

\noindent Now you need to directly output relations highly related to the following question and its subgoals in list format without other information or notes. \\
\noindent Q: \{\} \\
\noindent Subgoals: \{\} \\
\noindent Topic Entity: \{\} \\
\noindent Relations: \{\}
\end{tcolorbox}

\subsubsection{Entity Filtering}
\label{sec:entity_filtering_prompt}

\begin{tcolorbox}[
    colback=white, 
    colframe=black, 
    boxrule=0.5pt, 
    sharp corners,
    left=4pt,   
    right=4pt, 
    top=0pt,    
    bottom=2pt, 
    before skip=6pt,
    after skip=6pt
]
\small
\noindent Which entities in the following list ([] in Triples) can be used to answer question? Please provide the minimum possible number of entities, strictly following the constraints in the question. \\

\noindent \textit{In-Context Few-Shot} \\

\noindent Now you need to directly output the entities from [] in Triplets for the following question in list format without other information or notes. \\
\noindent Q: \{\} \\
\noindent Triplets: \{\}
\end{tcolorbox}

\subsubsection{Memory Update}
\label{sec:memory_update_prompt}

\begin{tcolorbox}[
    colback=white, 
    colframe=black, 
    boxrule=0.5pt, 
    sharp corners,
    left=4pt,   
    right=4pt, 
    top=0pt,    
    bottom=2pt, 
    before skip=6pt,
    after skip=6pt
]
\small
\noindent Based on the provided information (which may have missing parts and require further retrieval) and your own knowledge, output the currently known information required to achieve the subgoals. \\

\noindent \textit{In-Context Few-Shot} \\

\noindent Now you need to directly output the results of the following question in JSON format without other information or notes. \\
\noindent Q: \{\} \\
\noindent Subgoals: \{\} \\
\noindent Memory: \{\}
\end{tcolorbox}

\subsection{Answer Generation}
\label{sec:answer_generation_prompt}

\begin{tcolorbox}[
    colback=white, 
    colframe=black, 
    boxrule=0.5pt, 
    sharp corners,
    left=4pt,   
    right=4pt, 
    top=0pt,    
    bottom=2pt, 
    before skip=6pt,
    after skip=6pt
]
\small
\noindent Please answer the question based on the memory, related knowledge triplets and your knowledge. \\

\noindent \textit{In-Context Few-Shot} \\

\noindent Now you need to directly output the results of the following question in JSON format (must include "A" and "R") without other information or notes. If the triplets explicitly contains the answer to the question, prioritize the fact of the triplet over memory. \\
\noindent Q: \{\} \\
\noindent Memory: \{\} \\
\noindent Knowledge Triplets: \{\}
\end{tcolorbox}

\subsection{Failure-Aware Refinement}
\label{sec:failure_aware_refinement_prompts}

\subsubsection{Retrieval Necessity Diagnosis}
\label{sec:judge_reverse_prompt}

\begin{tcolorbox}[
    colback=white, 
    colframe=black, 
    boxrule=0.5pt, 
    sharp corners,
    left=4pt,   
    right=4pt, 
    top=0pt,    
    bottom=2pt, 
    before skip=6pt,
    after skip=6pt
]
\small
\noindent Based on the current frontier $\mathcal{E}_t$ and the historical evidence in memory $\mathcal{M}$, determine whether expanding the search space by adding previously pruned entities is necessary to resolve the reasoning impasse. \\

\noindent \textit{In-Context Few-Shot} \\

\noindent Now you need to directly output the results of the following question in the JSON format (must include "Add" and "Reason") without other information or notes. \\
\noindent Q: \{\} \\
\noindent Entities set to be retrieved: \{\} \\
\noindent Memory: \{\} \\
\noindent Knowledge Triplets: \{\}
\end{tcolorbox}

\subsubsection{Targeted Backtracking and Re-routing}
\label{sec:add_ent_prompt}

\begin{tcolorbox}[
    colback=white, 
    colframe=black, 
    boxrule=0.5pt, 
    sharp corners,
    left=4pt,   
    right=4pt, 
    top=0pt,    
    bottom=2pt, 
    before skip=6pt,
    after skip=6pt
]
\small
\noindent Under the guidance of memory $\mathcal{M}$ and current subgoals, select the minimum set of high-risk pruned entities to resume expansion and restore a verifiable evidence chain. \\

\noindent \textit{In-Context Few-Shot} \\

\noindent Now you need to directly output the results for the following Q in the list format without other information or notes. \\
\noindent Q: \{\} \\
\noindent Reason: \{\} \\
\noindent Candidate Entities: \{\} \\
\noindent Memory: \{\}
\end{tcolorbox}

\subsubsection{Grounded Inference}
\label{sec:grounded_inference_prompt}

\begin{tcolorbox}[
    colback=white, 
    colframe=black, 
    boxrule=0.5pt, 
    sharp corners,
    left=4pt,   
    right=4pt, 
    top=0pt,    
    bottom=2pt, 
    before skip=6pt,
    after skip=6pt
]
\small
\noindent Based on the verified path segments and unmet constraints, synthesize the most plausible answer. Prioritize entities in the [Graph Evidence], but leverage internal knowledge to bridge missing links and resolve entity identifiers. \\

\noindent \textit{In-Context Few-Shot} \\

\noindent Now you need to directly output the final answer string based on the following question and provided evidence. Do not output IDs or refusal phrases. \\
\noindent Question: \{\} \\
\noindent [Graph Evidence]: \{\} \\
\noindent Answer:
\end{tcolorbox}

\subsection{Blueprint Adaptation}
\label{sec:blueprint_adaptation_prompt}

\begin{tcolorbox}[
    colback=white, 
    colframe=black, 
    boxrule=0.5pt, 
    sharp corners,
    left=4pt,   
    right=4pt, 
    top=0pt,    
    bottom=2pt, 
    before skip=6pt,
    after skip=6pt
]
\small
\noindent Task: Generate or adapt a relational blueprint for the new question under the structural constraints of provided exemplars. \\

\noindent \textit{In-Context Few-Shot} \\

\noindent Now, identify the core semantic intent of the [New Question] and directly output its relational blueprint in list format. Use the exemplars only for structural reference. \\
\noindent Q: \{\} \\
\noindent Output:
\end{tcolorbox}

\section{Search SPARQL}
\label{sec:search_sparql}

To automatically process the KG data, we pre-define the SPARQL queries for Freebase, which can be executed by filling in the entity's \textit{mid} and relation.

\subsection{Relation Search}
\label{sec:relation_search}

\begin{sparqlbox}
\noindent \lstinline|PREFIX ns: <http://rdf.freebase.com/ns/>| \\
\noindent \lstinline|SELECT DISTINCT ?relation| \\
\noindent \lstinline|WHERE {| \\
\noindent \hspace*{1em} \lstinline|ns:mid ?relation ?x .| \\
\noindent \lstinline|}|
\end{sparqlbox}

\begin{sparqlbox}
\noindent \lstinline|PREFIX ns: <http://rdf.freebase.com/ns/>| \\
\noindent \lstinline|SELECT DISTINCT ?relation| \\
\noindent \lstinline|WHERE {| \\
\noindent \hspace*{1em} \lstinline|?x ?relation ns:mid .| \\
\noindent \lstinline|}|
\end{sparqlbox}

\subsection{Entity Search}
\label{sec:entity_search}

\begin{sparqlbox}
\noindent \lstinline|PREFIX ns: <http://rdf.freebase.com/ns/>| \\
\noindent \lstinline|SELECT ?tailEntity| \\
\noindent \lstinline|WHERE {| \\
\noindent \hspace*{1em} \lstinline|ns:mid ns:relation ?tailEntity .| \\
\noindent \lstinline|}|
\end{sparqlbox}

\begin{sparqlbox}
\noindent \lstinline|PREFIX ns: <http://rdf.freebase.com/ns/>| \\
\noindent \lstinline|SELECT ?tailEntity| \\
\noindent \lstinline|WHERE {| \\
\noindent \hspace*{1em} \lstinline|?tailEntity ns:relation ns:mid .| \\
\noindent \lstinline|}|
\end{sparqlbox}

\subsection{Entity ID Resolution}
\label{sec:entity_id_resolution}

\begin{sparqlbox}
\begin{lstlisting}[language=SPARQL]
PREFIX ns: <http://rdf.freebase.com/ns/>
SELECT DISTINCT ?tailEntity
WHERE {
  BIND(ns:id AS ?entity)
  {
    ?entity ns:type.object.name ?tailEntity .
    FILTER(LANG(?tailEntity) = "" || LANGMATCHES(LANG(?tailEntity), "en"))
    BIND(1 AS ?priority)
  }
  UNION
  {
    ?entity ns:common.topic.alias ?tailEntity .
    FILTER(LANG(?tailEntity) = "" || LANGMATCHES(LANG(?tailEntity), "en"))
    BIND(2 AS ?priority)
  }
  UNION
  {
    ?entity <http://www.w3.org/2002/07/owl#sameAs> ?tailEntity .
    BIND(3 AS ?priority)
  }
}
ORDER BY ASC(?priority) LIMIT 1
\end{lstlisting}
\end{sparqlbox}

\section{Datasets}
\label{sec:datasets}

We evaluate our method on three complex KGQA benchmarks: \textbf{WebQSP}~\citep{webqsp}
, \textbf{ComplexWebQuestions (CWQ)}~\citep{cwq}, and \textbf{GrailQA}~\citep{grailqa}. All datasets are constructed on the Freebase knowledge graph~\citep{freebase}. The detailed statistics are summarized in Table~\ref{tab:dataset_statistics}.

\paragraph{WebQSP} 
WebQSP consists of 4,737 natural language questions that require 1-hop or 2-hop reasoning over Freebase. It is widely used to evaluate the robustness of Entity Linking and multi-hop reasoning capabilities.

\paragraph{ComplexWebQuestions (CWQ)} 
CWQ extends WebQSP by introducing four types of complex constraints: \textit{conjunction}, \textit{composition}, \textit{comparative}, and \textit{superlative}. It contains 34,689 questions requiring multi-hop reasoning (up to 4 hops) and strict constraint handling.

\paragraph{GrailQA} 
GrailQA is a large-scale dataset with 64,331 questions designed to test three levels of generalization: \textit{I.I.D.}, \textit{Compositional}, and \textit{Zero-Shot}. It poses significant challenges for models to handle novel schemas and diverse logical structures not seen during training.

\begin{table}[h!]
    \centering
    \resizebox{\columnwidth}{!}{%
        \begin{tabular}{lcccc}
            \toprule
            \textbf{Dataset} & \textbf{Answer Format} & \textbf{Train} & \textbf{Test} & \textbf{Licence} \\
            \midrule
            ComplexWebQuestions & Entity & 27,734 & 3,531 & - \\
            WebQSP & Entity/Number & 3,098 & 1,639 & CC Licence \\
            GrailQA & Entity/Number & 44,337 & 1,000 & - \\
            \bottomrule
        \end{tabular}%
    }
    \caption{Statistics of KGQA datasets.}
    \label{tab:dataset_statistics} 
\end{table}

\section{Baseline Descriptions}
\label{sec:baseline_descriptions}

To comprehensively evaluate our method, we compare it against two categories of state-of-the-art approaches: (1) LLM-only methods without external knowledge access; and (2) KG-augmented LLM methods, which include both fine-tuned models and training-free prompting frameworks.

\subsection{LLM-Only Methods}
\begin{itemize}[leftmargin=*, nosep]
    \item \textbf{Standard Prompting (IO Prompt)}~\citep{io} instructs the LLM to generate answers directly based on the input question, establishing a baseline for the model's intrinsic knowledge capacity in a few-shot setting.
    \item \textbf{Chain-of-Thought (CoT)}~\citep{cot} prompts LLMs to generate a sequence of intermediate reasoning steps before deriving the final answer, rather than outputting the result directly.
    \item \textbf{Self-Consistency (SC)}~\citep{sc} samples multiple diverse reasoning paths via CoT and aggregates the results through majority voting to improve answer stability and accuracy.
\end{itemize}

\subsection{Fine-tuned KG-Augmented LLM Methods}
\begin{itemize}[leftmargin=*, nosep]
    \item \textbf{RE-KBQA}~\citep{rekbqa} incorporates additional supervision signals to emphasize relation exploration and improve the representation of KG entities for reasoning path selection.
    \item \textbf{UniKGQA}~\citep{unikgqa} integrates the retrieval and reasoning modules into a single shared model architecture to facilitate semantic alignment between questions and KG entities.
    \item \textbf{RoG}~\citep{rog} generates faithful reasoning plans grounded in the graph structure to guide LLM decoding, thereby improving interpretability and reasoning reliability.
    \item \textbf{DeCAF}~\citep{decaf} combines semantic parsing with LLMs to jointly decode valid logical forms and natural language answers.
    \item \textbf{KG-Agent}~\citep{kgagent} formalizes multi-hop reasoning as an executable program, incorporating KG querying tools and memory mechanisms to enable step-by-step logical derivation.
    \item \textbf{RnG-KBQA}~\citep{rngkbqa} employs a generate-and-rank paradigm where candidate logical programs are first enumerated and ranked by BERT, and then refined into more complex structures using T5.
    \item \textbf{TIARA}~\citep{tiara} adopts a multi-grained retrieval strategy using BERT for schema linking, followed by a T5-based generation module that outputs constrained logical plans.
    \item \textbf{FC-KBQA}~\citep{fckbqa} utilizes a fine-to-coarse composition strategy to decouple knowledge acquisition from reasoning, aiming for better generalization across diverse KG schemas.
    \item \textbf{Pangu}~\citep{pangu} develops an end-to-end KBQA architecture that focuses on compositional generalization to handle complex query structures.
    \item \textbf{FlexKBQA}~\citep{flexkbqa} proposes a few-shot framework designed to adapt to new KGs and query languages using a limited number of annotated samples.
    \item \textbf{GAIN}~\citep{gain} introduces a specialized data augmentation mechanism for KGQA tasks to improve model robustness against out-of-distribution shifts.
\end{itemize}

\subsection{Prompting KG-Augmented LLM Methods}
\begin{itemize}[leftmargin=*, nosep]
    \item \textbf{KB-BINDER}~\citep{kbbinder} relies on few-shot in-context learning to map natural language queries into executable logical forms, explicitly aligning question semantics with underlying schema items.
    \item \textbf{KD-CoT}~\citep{Kd} augments standard Chain-of-Thought prompting by interleaving KG retrieval, utilizing external evidence at each reasoning step to ground the LLM's ongoing logical trace.
    \item \textbf{StructGPT}~\citep{structgpt} casts structured data reasoning as a tool-use problem, equipping the LLM with specialized programmatic interfaces to interactively query and filter information from the graph.
    \item \textbf{ToG}~\citep{tog} formulates the reasoning process as an LLM-guided beam search, where the agent dynamically evaluates and expands candidate entity relations to construct optimal evidence paths.
    \item \textbf{PoG}~\citep{pog} directs online graph exploration by initially generating a sequence of subgoals, while integrating historical context tracing and self-correction modules to maintain logical consistency throughout the multi-hop retrieval.
\end{itemize}

\section{Implementation Details}
\label{sec:appendix_details}

\subsection{Experiment Settings}
\label{sec:exp_settings}
In our experiments, we evaluate \textbf{CoG} using three different language models: GPT-3.5 and GPT-4 accessed via the OpenAI API \footnote{\url{https://platform.openai.com/docs}}, and Qwen2.5  \footnote{\url{https://huggingface.co/Qwen}}. We set the temperature parameter to 0.3, frequency penalty to 0, and presence penalty to 0. The maximum token length for generation is 1024. In all experiments, the depth of exploration is set to 4 to avoid endless exploration. The experiments are conducted on a Linux server equipped with two Intel(R) Xeon(R) Gold 6148 CPUs @ 2.40GHz and 256 GB RAM.

\subsection{Blueprint Construction and Copy–Adapt Analysis}
\label{sec:blueprint_construction_sensitivity}

This section details the offline relational blueprint template library construction and provides an empirical analysis of how the copy threshold $\tau_{copy}$ regulates the \textbf{copy–adapt mechanism}—the trade-off between utilizing structural priors and maintaining generative flexibility. For semantic retrieval, the anchors of these templates are encoded using \texttt{msmarco-distilbert-base-tas-b}\footnote{\url{https://huggingface.co/sentence-transformers/msmarco-distilbert-base-tas-b}}. Our sensitivity experiments are conducted on the WebQSP dataset using GPT-4o-mini.

\subsubsection{Blueprint Library Statistics}
\label{sec:Blueprint Library Statistics}
We construct dataset-specific relational blueprint template libraries by distilling structural patterns from training splits. Table~\ref{tab:blueprint_stats} summarizes the statistics, where a defining characteristic is the extreme \textbf{structural compression}. For instance, on the large-scale GrailQA dataset, over 44,000 training questions condense into merely 3,703 unique blueprint templates—a ratio of only 8.3\%. 

This significant reduction empirically validates that the logical topologies of KG reasoning are far more finite and recurrent than their natural language surface forms. By abstracting away concrete entities, the library captures generic reasoning prototypes (e.g., multi-hop filtration, comparative logic) that are highly transferable to unseen queries.

\begin{table}[h]
    \centering
    \footnotesize
    \renewcommand{\arraystretch}{1.1}
    \setlength{\tabcolsep}{2pt}
    \begin{tabular*}{\columnwidth}{@{\extracolsep{\fill}}lccc}
        \toprule
        \textbf{Dataset} & \textbf{Train Size} & \textbf{Lib Size} & \textbf{Ratio} \\
         & \textbf{(Queries)} & \textbf{(Templates)} & \textbf{(\%)} \\
        \midrule
        \textbf{WebQSP}  & 3,098  & 569   & 18.4\% \\
        \textbf{CWQ}     & 27,734 & 6,747 & 24.3\% \\
        \textbf{GrailQA} & 44,337 & 3,703 & 8.3\% \\
        \bottomrule
    \end{tabular*}
    \caption{Statistics of offline relational blueprint libraries. The compression ratio highlights the high reusability of structural patterns.}
    \label{tab:blueprint_stats}
\end{table}

\subsubsection{Sensitivity of Copy Threshold}
\label{sec:Sensitivity_of_Copy_Threshold}
The threshold $\tau_{copy}$ acts as the gatekeeper of the \textbf{copy--adapt strategy}, determining when to trust structural priors (copy) versus when to rely on LLM generalization (adapt) . Table~\ref{tab:threshold_sensitivity} presents the impact of $\tau_{copy}$ on performance, revealing an ``inverted-U'' trajectory where the optimal accuracy (83.7\%) is achieved at $\tau_{copy}=0.92$. Crucially, we observe a robust performance plateau within the $[0.9, 1.0]$ interval, where Hits@1 consistently exceeds 82\%, indicating that CoG is not hypersensitive to specific threshold tuning.

\paragraph{Balancing copy and adapt.}
A potential concern regarding methods utilizing historical data is whether relying on training-derived blueprints limits generalization compared to pure zero-shot approaches like ToG or GoG. The strategy distribution at the optimal threshold ($\tau_{copy}=0.92$) provides compelling counter-evidence:
\begin{itemize}
    \item \textbf{Minimal Dependency, Maximum Gain:} At peak performance, the model invokes the copy mode for only \textbf{8.7\%} of queries (high-confidence matches), while adapting via LLM generation for the remaining \textbf{91.3\%}. This confirms that CoG does not rigidly overfit to training distributions.
    \item \textbf{Dynamic Fallback:} The blueprint library serves as a ``compass'' for frequent patterns. For novel or unseen structures (similarity $< \tau_{copy}$), CoG seamlessly transitions to its adapt phase, retaining the full generative flexibility of zero-shot baselines while benefiting from the stability of structural priors when applicable .
\end{itemize}

\begin{table}[h]
    \centering
    \footnotesize
    \renewcommand{\arraystretch}{1.2}
    \setlength{\tabcolsep}{1pt}
    \begin{tabular*}{\columnwidth}{@{\extracolsep{\fill}}c cc c}
        \toprule
        \textbf{Threshold} & \multicolumn{2}{c}{\textbf{Copy--Adapt Strategy (Queries)}} & \textbf{Hits@1} \\
        \cmidrule{2-3}
        ($\tau_{copy}$) & \textbf{Copy (Memory)} & \textbf{Adapt (Gen.)} & \textbf{(\%)} \\
        \midrule
        0.70 & 1639 (100\%) & 0 & 78.9 \\
        0.80 & 1355 (82.7\%) & 284 & 80.6 \\
        0.90 & 193 (11.8\%) & 1446 & 82.8 \\
        \textbf{0.92} & \textbf{143 (8.7\%)} & \textbf{1496} & \textbf{83.7} \\
        0.95 & 74 (4.5\%) & 1565 & 82.2 \\
        1.00 & 0 (0.0\%) & 1639 & 82.7 \\
        \bottomrule
    \end{tabular*}
    \caption{Impact of the copy threshold $\tau_{copy}$ on model behavior. Optimal performance balances selective memory usage with generative fallback.}
    \label{tab:threshold_sensitivity}
\end{table}

\subsection{Sensitivity Analysis of Reranking Weights}
\label{sec:sensitivity_reranking}

In our main experiments, we employ \texttt{gpt-4o-mini} as the backbone agent for reasoning and scoring. The final reranking score is a weighted sum of three components: local semantic relevance ($\lambda_{loc}$), step-wise structural alignment ($\lambda_{step}$), and global compatibility ($\lambda_{glob}$), subject to $\sum \lambda = 1$.
In this section, we provide a detailed sensitivity analysis to justify our hyperparameter selection ($\lambda_{loc}=0.6, \lambda_{step}=0.25, \lambda_{glob}=0.15$) and demonstrate the robustness of our approach.

\subsubsection{Experimental Design and Rationale}
To effectively decouple the impact of semantic signals versus structural constraints, we designed a two-stage sensitivity test:

\paragraph{(1) Main Trend: Balancing Semantics vs. Structure.}
\begin{itemize}
    \item \textbf{Setup:} We vary the primary weight $\lambda_{loc}$ from 0.4 to 0.8. Crucially, while adjusting $\lambda_{loc}$, we maintain a fixed ratio between the two structural weights ($\lambda_{step} : \lambda_{glob} \approx 5:3$).
    \item \textbf{Rationale:} This setup treats "structural constraint" as a unified force. By adjusting $\lambda_{loc}$, we control the trade-off between \textit{trusting the LLM's local semantic matching} (high $\lambda_{loc}$) versus \textit{trusting the Blueprint's structural guidance} (low $\lambda_{loc}$).
\end{itemize}

\paragraph{(2) Internal Ratio Variants: The Composition of Structure.}
\begin{itemize}
    \item \textbf{Setup:} We fix $\lambda_{loc}$ at its optimal value ($0.6$) and disturb the internal distribution of the remaining $0.4$ weight mass. We test variants where the structural weight is dominated by global compatibility (e.g., $\lambda_{glob}=0.2, 0.3$) or equally distributed.
    \item \textbf{Rationale:} We hypothesize that \textit{step-wise alignment} (checking the validity of each reasoning hop) provides a more precise signal than \textit{global compatibility} (which only checks the final path shape). Therefore, $\lambda_{step}$ should logically be assigned a higher weight than $\lambda_{glob}$.
\end{itemize}

\paragraph{(3) Uniform Baseline.}
We also compare against a naive baseline where $\lambda_{loc}=\lambda_{step}=\lambda_{glob}\approx1/3$, to verify that the performance gains stem from our specific weighting strategy rather than simple score ensembling.

\subsubsection{Results and Analysis}
\begin{figure}[h]
    \centering
    \includegraphics[width=1.0\columnwidth]{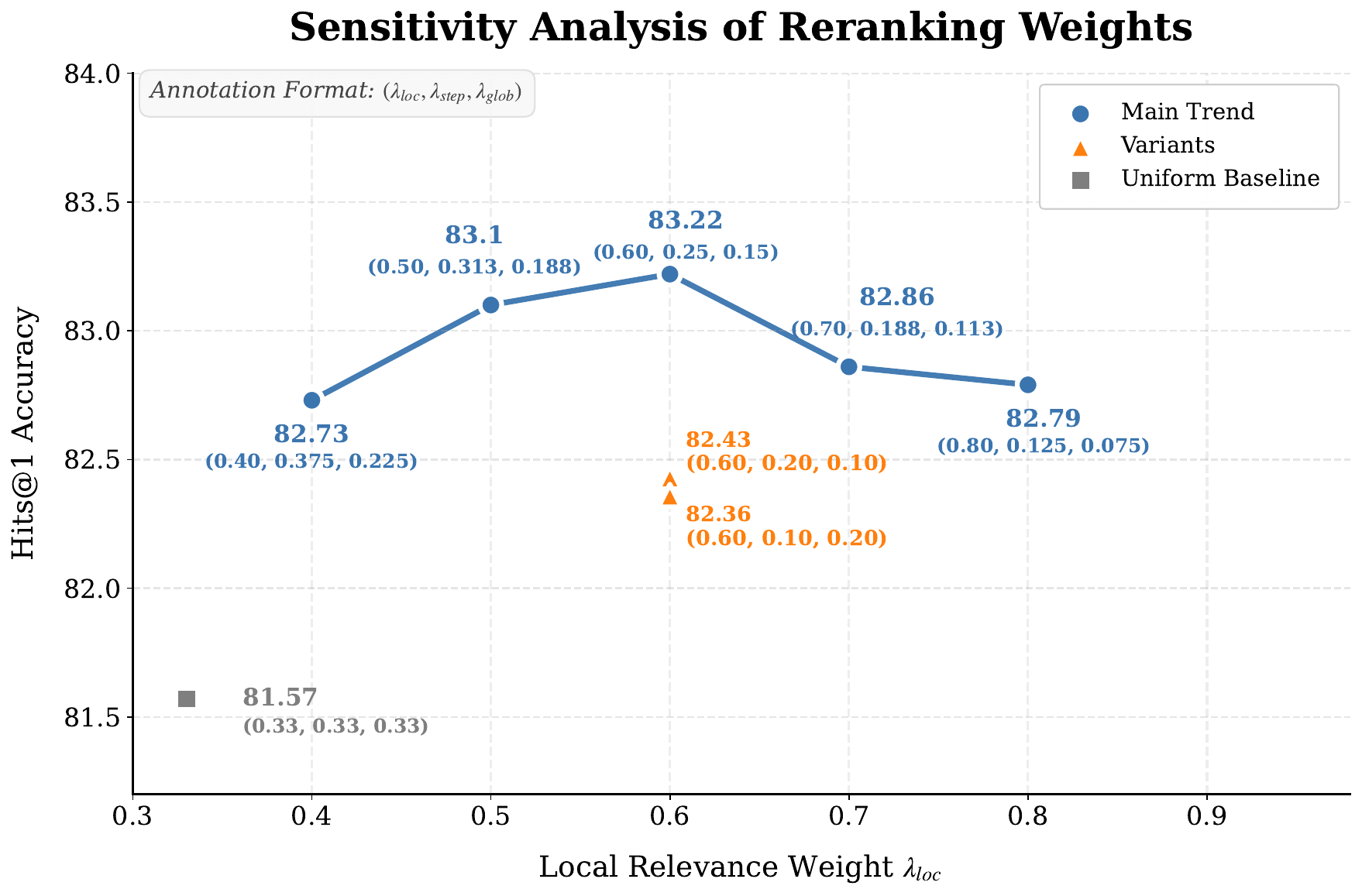} 
    \caption{Sensitivity analysis of reranking weights. The \textcolor{blue}{blue line} tracks performance as $\lambda_{loc}$ varies. The \textbf{\textcolor{orange}{orange triangles}} represent variants with suboptimal internal structural ratios at the peak $\lambda_{loc}=0.6$. The \textbf{\textcolor{gray}{grey square}} denotes the naive uniform baseline. The explicit weight configuration $(\lambda_{loc}, \lambda_{step}, \lambda_{glob})$ is annotated for each point.}
    \label{fig:sensitivity_appendix}
\end{figure}

The results on WebQSP are visualized in Figure~\ref{fig:sensitivity_appendix}.

\paragraph{Optimal Balance Found at $\lambda_{loc}=0.6$.}
The blue curve exhibits a clear inverted-U trajectory.
\begin{itemize}
    \item \textbf{Under-weighting Semantics ($\lambda_{loc} < 0.6$):} When $\lambda_{loc}$ is low (e.g., 0.4), the strict structural constraints may filter out semantically correct entities that have weaker structural signals (e.g., due to sparse KG connections), leading to a performance drop (82.73\%).
    \item \textbf{Over-weighting Semantics ($\lambda_{loc} > 0.6$):} As $\lambda_{loc}$ approaches 0.8, the agent behaves like a greedy semantic searcher, ignoring global blueprint constraints, which drops accuracy to 82.79\%.
\end{itemize}

\paragraph{Superiority of Step-wise Verification.}
The comparison at $\lambda_{loc}=0.6$ (orange markers) strongly supports our hypothesis regarding internal structural ratios.
Even with the optimal primary weight, allocating more weight to global compatibility ($\lambda_{glob} > \lambda_{step}$) leads to significant degradation (82.43\% and 82.36\%). This confirms that \texttt{gpt-4o-mini} benefits more from fine-grained, step-by-step structural verification than from coarse-grained global checks.

\section{Detailed Performance Breakdown}
\label{app:breakdown}

\begin{figure}[h]
    \centering
    \includegraphics[width=0.95\columnwidth]{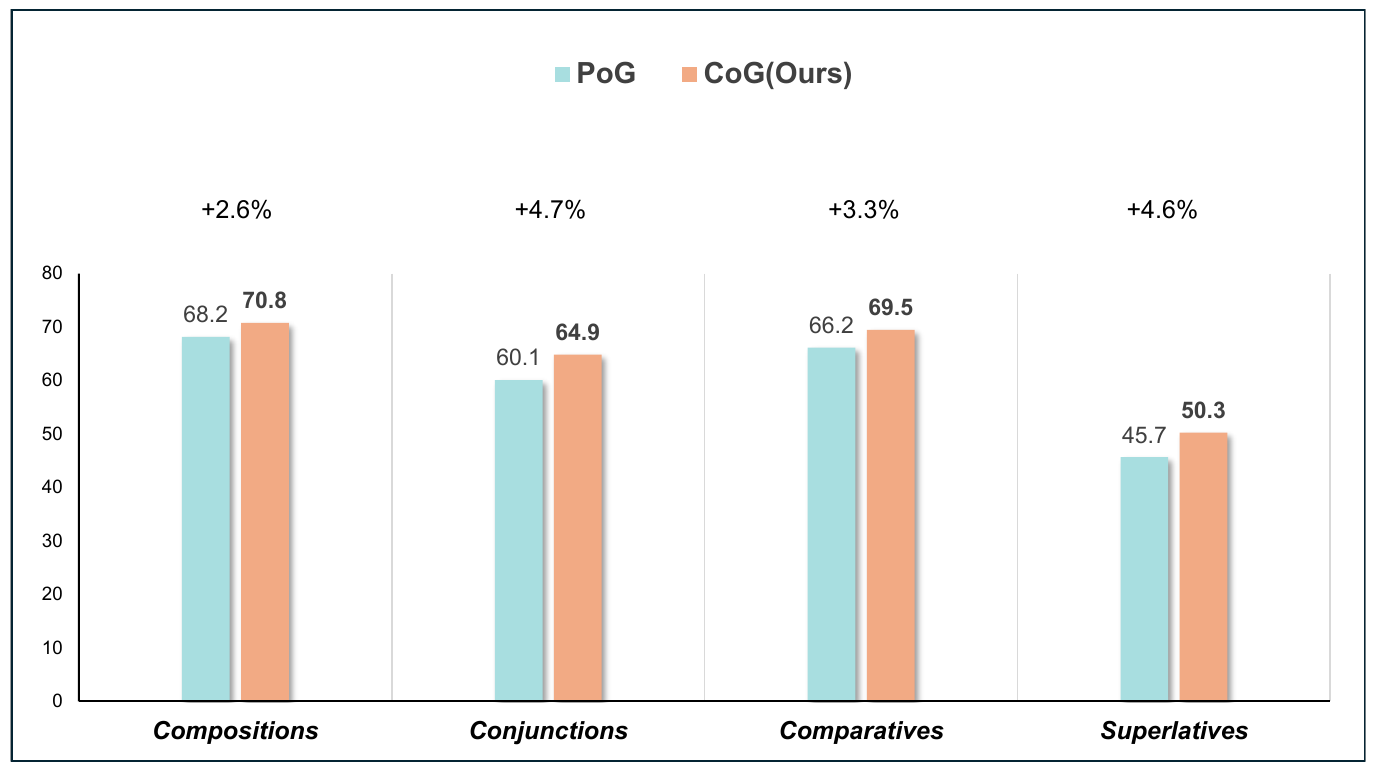} 
    \caption{Performance breakdown by query type. Our method achieves significant accuracy gains, particularly in structurally complex categories like \textbf{Conjunction (+4.7\%)} and \textbf{Superlative (+4.6\%)}.}
    \label{fig:breakdown}
\end{figure}

Figure~\ref{fig:breakdown} details the performance comparison across four query types on CWQ. Our method (CoG) consistently outperforms the baseline across all categories, with the performance gap widening as structural complexity increases.

Most notably, we observe the substantial accuracy gains in \textbf{Conjunction (+4.7\%)} and \textbf{Superlative (+4.6\%)} queries. These categories demand rigorous logic: conjunctions require satisfying multi-branch constraints simultaneously, while superlatives necessitate global comparison over candidate sets. The significant boost here substantiates that our blueprint-guided constraints effectively prevent reasoning drift in complex scenarios. Meanwhile, performance on standard \textbf{Composition (+2.6\%)} and \textbf{Comparative (+3.3\%)} chains remains robust, confirming that our structural constraints enhance reliability without compromising efficiency on fundamental tasks.

\section{Analysis of Failure-Aware Refinement}
\label{app:refinement}

\begin{figure}[h]
    \centering
    \includegraphics[width=1.0\columnwidth]{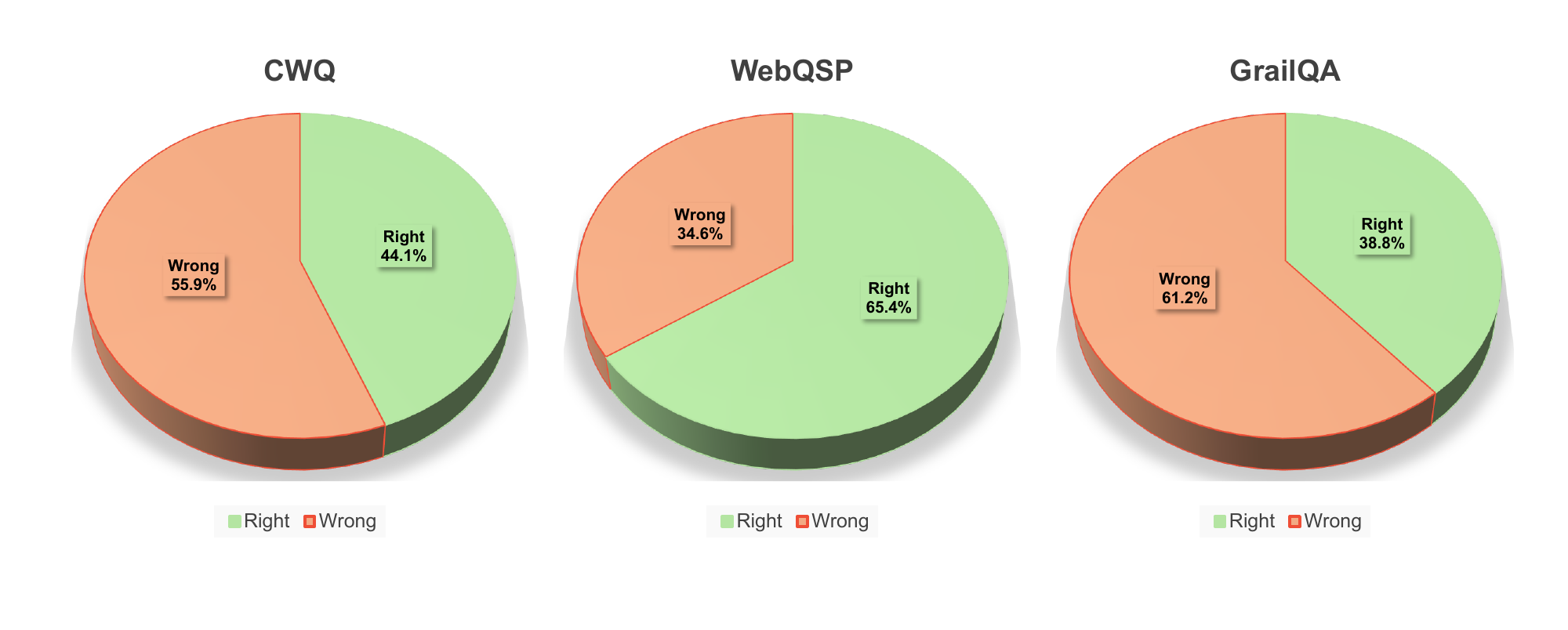} 
    \caption{Outcome distribution of the refinement process. The "Right" segment represents the proportion of queries successfully rectified by the module after being triggered.}
    \label{fig:refinement_pie}
\end{figure}

We scrutinize the activation frequency and error-correction capability of the \textit{Failure-Aware Refinement} module across datasets. Table~\ref{tab:refinement_stats} details the triggering statistics, while Figure~\ref{fig:refinement_pie} breaks down the outcomes of these triggered cases.

\paragraph{Activation and Complexity Correlation.}
Table~\ref{tab:refinement_stats} reveals a clear correlation between query complexity and refinement activation. The module is triggered most frequently on \textbf{CWQ (45.3\%)}, significantly higher than on WebQSP (32.3\%) and GrailQA (29.9\%). This high activation rate on CWQ reflects the inherent difficulty of its multi-hop questions, where intermediate reasoning steps are prone to "dead ends," thereby necessitating frequent intervention by our safety-net mechanism.

\paragraph{Recovery Effectiveness.}
Figure~\ref{fig:refinement_pie} illustrates the success rate of the refinement module in rectifying potential failures.
\begin{itemize}
    \item \textbf{I.I.D. vs. Zero-Shot:} The module proves most effective on the I.I.D. WebQSP dataset, recovering \textbf{65.4\%} of triggered queries. In contrast, performance dips on the zero-shot GrailQA dataset (38.8\%). This variance suggests that while self-correction is powerful for standard reasoning, it faces challenges when grounding novel entities without prior schema exposure.
    \item \textbf{Contribution to Robustness:} Despite the lower recovery rate on complex datasets, rectifying nearly \textbf{40\%} of potential failures on CWQ and GrailQA constitutes a critical portion of the overall performance gains reported in the main results.
\end{itemize}

\begin{table}[h]
    \centering
    \small
    \renewcommand{\arraystretch}{1.1} 
    \setlength{\tabcolsep}{5pt}
    \begin{tabular}{l c c c c}
        \toprule
        \multirow{2}{*}{\textbf{Dataset}} & \textbf{Total} & \textbf{Overall} & \multicolumn{2}{c}{\textbf{Refinement Trigger}} \\
        \cmidrule(lr){4-5}
         & \textbf{Queries} & \textbf{Hits@1} & \textbf{Count} & \textbf{Rate} \\
        \midrule
        \textbf{CWQ}     & 3531 & 66.9 & 1599 & 45.3\% \\
        \textbf{WebQSP}  & 1639 & 86.8 & 529  & 32.3\% \\
        \textbf{GrailQA} & 1000 & 79.2 & 299  & 29.9\% \\
        \bottomrule
    \end{tabular}
    \caption{Statistics of refinement activation. We report the frequency (\textbf{Count}) and proportion (\textbf{Rate}) of queries necessitating self-correction intervention.}
    \label{tab:refinement_stats}
\end{table}

\section{Quantitative Error Analysis}
\label{sec:error_analysis}

\begin{figure}[t]
    \centering
    \includegraphics[width=0.95\linewidth]{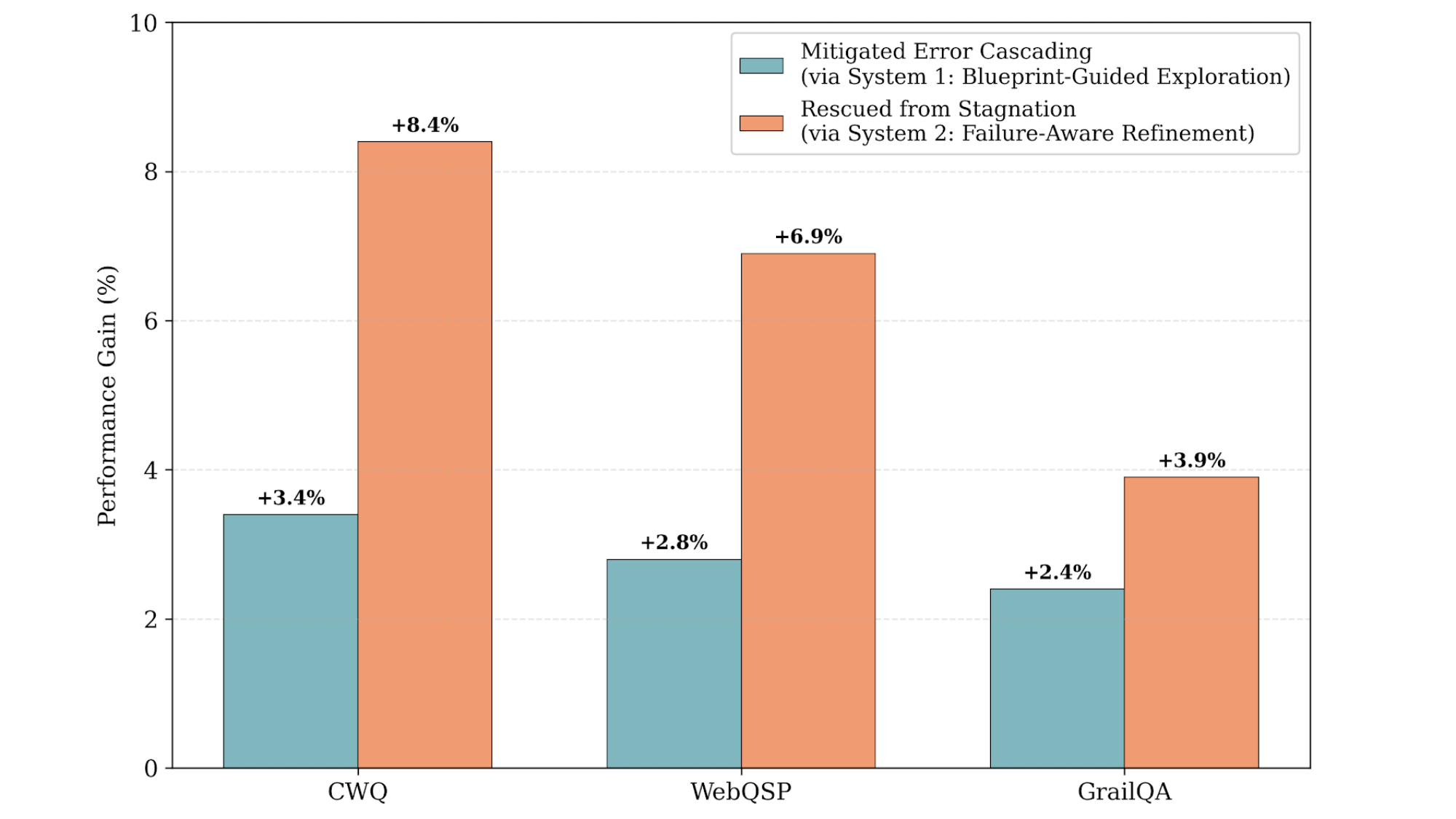}
    \caption{Quantitative attribution of performance gains to CoG's core mechanisms. The \textbf{blue bars} quantify the mitigation of \textit{Error Cascading} via \textbf{System 1} (Blueprint-Guided Exploration), acting as a structural filter against noise. The \textbf{orange bars} represent the recovery from \textit{Reasoning Stagnation} via \textbf{System 2} (Failure-Aware Refinement), resolving structural misalignment through diagnostic backtracking. The pronounced impact on CWQ highlights the necessity of the refinement mechanism for complex multi-hop reasoning.}
    \label{fig:error_analysis}
\end{figure}

To explicitly quantify how CoG mitigates the two core dimensions of cognitive rigidity—\textit{Error Cascading} and \textit{Reasoning Stagnation}—we conducted an ablation-based attribution analysis. We define the performance gap between the full CoG framework and its component-ablated variants as a proxy for the volume of specific errors corrected by our mechanisms. Figure~\ref{fig:error_analysis} visualizes these gains across three benchmarks.

\vspace{0.3em}
\noindent\textbf{Mitigating Error Cascading (Blue Bars).} 
The blue bars in Figure~\ref{fig:error_analysis} quantify the contribution of \textit{Blueprint-Guided Exploration} (System 1). By enforcing structural constraints, CoG consistently filters out neighborhood noise that typically misleads baselines into irreversible deviations. On the noise-intensive CWQ dataset, this mechanism rescues \textbf{3.4\%} of queries from cascading errors. Even on simpler datasets like WebQSP and GrailQA, the consistent gains (+2.8\% and +2.4\%) confirm that structural blueprints serve as a robust, universal filter against the "butterfly effect" of early selection errors.

\vspace{0.3em}
\noindent\textbf{Resolving Stagnation (Orange Bars).} 
The orange bars highlight the critical role of \textit{Failure-Aware Refinement} (System 2) in overcoming reasoning dead-ends. This effect is particularly dominant on the complex CWQ benchmark, where the diagnostic re-routing mechanism recovers a substantial \textbf{8.4\%} of queries that initially stagnated. The sharp contrast between the high gain on CWQ and the moderate gains on other datasets suggests a clear correlation: as reasoning depth and complexity increase, the risk of structural misalignment grows, making CoG's "safety net" mechanism increasingly indispensable for navigating treacherous search spaces.

\section{Generalization Analysis on Heterogeneous Knowledge Graphs}
\label{app:generalization}

To assess the robustness of CoG beyond the specific schema of Freebase, we extended our evaluation to \textbf{Wikidata}~\citep{wikidata}, a knowledge graph characterized by significantly larger scale and greater heterogeneity. This setting serves as a rigorous stress test for the model's ability to handle schema variations and navigate noisy environments without specific fine-tuning.

\paragraph{Experimental Setup.}
We mapped the entity annotations in WebQSP and CWQ from Freebase IDs (MIDs) to Wikidata IDs (QIDs). It is important to acknowledge that due to ontological misalignment between the two KGs, a subset of entities could not be perfectly mapped, introducing an inherent layer of noise and information loss. We benchmarked CoG against the strongest baselines, ToG and PoG, under this challenging setting.

\paragraph{Results and Discussion.}
The comparative results are summarized in Table~\ref{tab:wiki_results}. We first observe a universal performance decline across all methods when transitioning from Freebase to Wikidata. This drop is anticipated and stems primarily from two factors: (1) the \textit{annotation bias}, as the datasets were originally curated for Freebase, rendering the mapping process lossy; and (2) the \textit{structural complexity}, where Wikidata's dense topology substantially expands the search space and complicates relation filtering.

\begin{table}[t]
\centering
\small
\renewcommand{\arraystretch}{1.2}
\begin{tabular}{l|cc}
\toprule
\textbf{Method} & \textbf{WebQSP} & \textbf{CWQ} \\
\midrule
\multicolumn{3}{l}{\textit{Source Domain: Freebase}} \\
\midrule
ToG & 76.2 & 57.1 \\
PoG & 82.0 & 63.2 \\
\textbf{CoG (Ours)} & \textbf{86.8} & \textbf{66.9} \\
\midrule
\multicolumn{3}{l}{\textit{Target Domain: Wikidata}} \\
\midrule
ToG & 68.6 & 54.9 \\
PoG & 73.8 & 60.7 \\
\textbf{CoG (Ours)} & \textbf{76.5} & \textbf{62.8} \\
\bottomrule
\end{tabular}
\caption{Performance comparison using different source KGs (Freebase vs. Wikidata). Despite the domain shift, CoG demonstrates superior robustness against schema heterogeneity compared to ToG and PoG.}
\label{tab:wiki_results}
\end{table}

Despite these environmental hurdles, CoG maintains a distinct advantage over the baselines. Specifically, CoG outperforms the strongest competitor, PoG, by margins of \textbf{2.7\%} on WebQSP and \textbf{2.1\%} on CWQ. These results offer two critical insights into the mechanism of CoG:
\begin{itemize}
    \item \textbf{Schema Agnosticism:} The Relational Blueprint mechanism appears to capture abstract reasoning patterns (e.g., \textit{Subject} $\to$ \textit{Attribute} $\to$ \textit{Value}) rather than overfitting to specific naming conventions (e.g., Freebase's \texttt{people.person.place\_of\_birth}). This abstraction allows the generated blueprints to adapt effectively to the distinct property predicates found in Wikidata.
    \item \textbf{Resilience to Noise:} The performance gap suggests that the \textit{Failure-Aware Refinement} module is particularly effective in heterogeneous settings. When initial search paths are misled by Wikidata's noise, CoG's diagnostic backtracking prevents the reasoning chain from premature collapse—a failure mode frequently observed in baselines lacking such corrective mechanisms.
\end{itemize}

\section{Case Studies}
\label{sec:case}
We conduct a qualitative analysis of two representative cases to scrutinize the interplay between structural priors and reasoning stability. 

\begin{figure*}[htbp]
    \centering
    \includegraphics[width=0.95\textwidth]{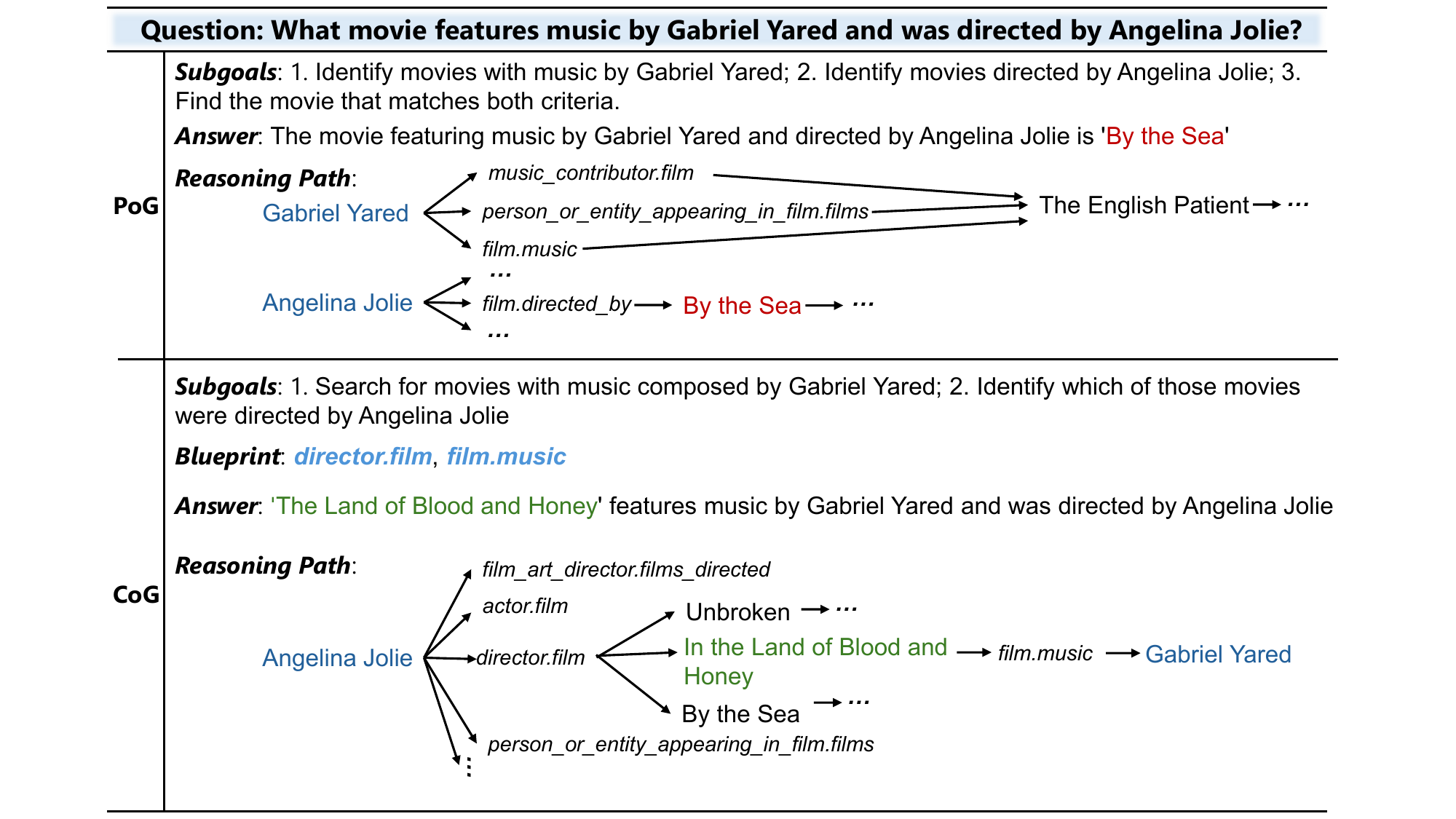}
    \caption{Case 1: Comparison of reasoning trajectories on a super-node. PoG is misled by the local density of acting roles, leading to semantic drift, while CoG utilizes the relational blueprint to filter distractors and maintain structural consistency.}
    \label{fig:case1}
\end{figure*}

\noindent\textbf{Case 1} (Figure~\ref{fig:case1}) exemplifies CoG's capacity to navigate the ``neighborhood noise'' inherent in super-nodes, where local semantic cues often diverge from the underlying reasoning logic. In this query concerning Angelina Jolie—an entity with an overwhelming density of acting-related metadata—the baseline PoG, lacking global structural constraints, succumbs to the high-frequency semantic traps of her acting career. Consequently, PoG exhausts 5,544 tokens drifting through unrelated films (e.g., \textit{The English Patient}) before eventually hallucinating \textit{By the Sea} as a partially correct but logically invalid answer. In contrast, CoG strictly adheres to the adapted relational blueprint $\langle \text{film.director.film, film.film.music} \rangle$. Even though the local semantic score for the ``actor'' relation is dominant, the step-wise alignment signal $\phi_{step}$ effectively penalizes these noisy distractors. By anchoring the search within the directorial slot, CoG identifies the correct answer, \textit{In the Land of Blood and Honey}, with minimal overhead, demonstrating that relational blueprints function as a robust ``structural compass'' in complex semantic landscapes.

\begin{figure*}[htbp]
    \centering
    \includegraphics[width=0.95\textwidth]{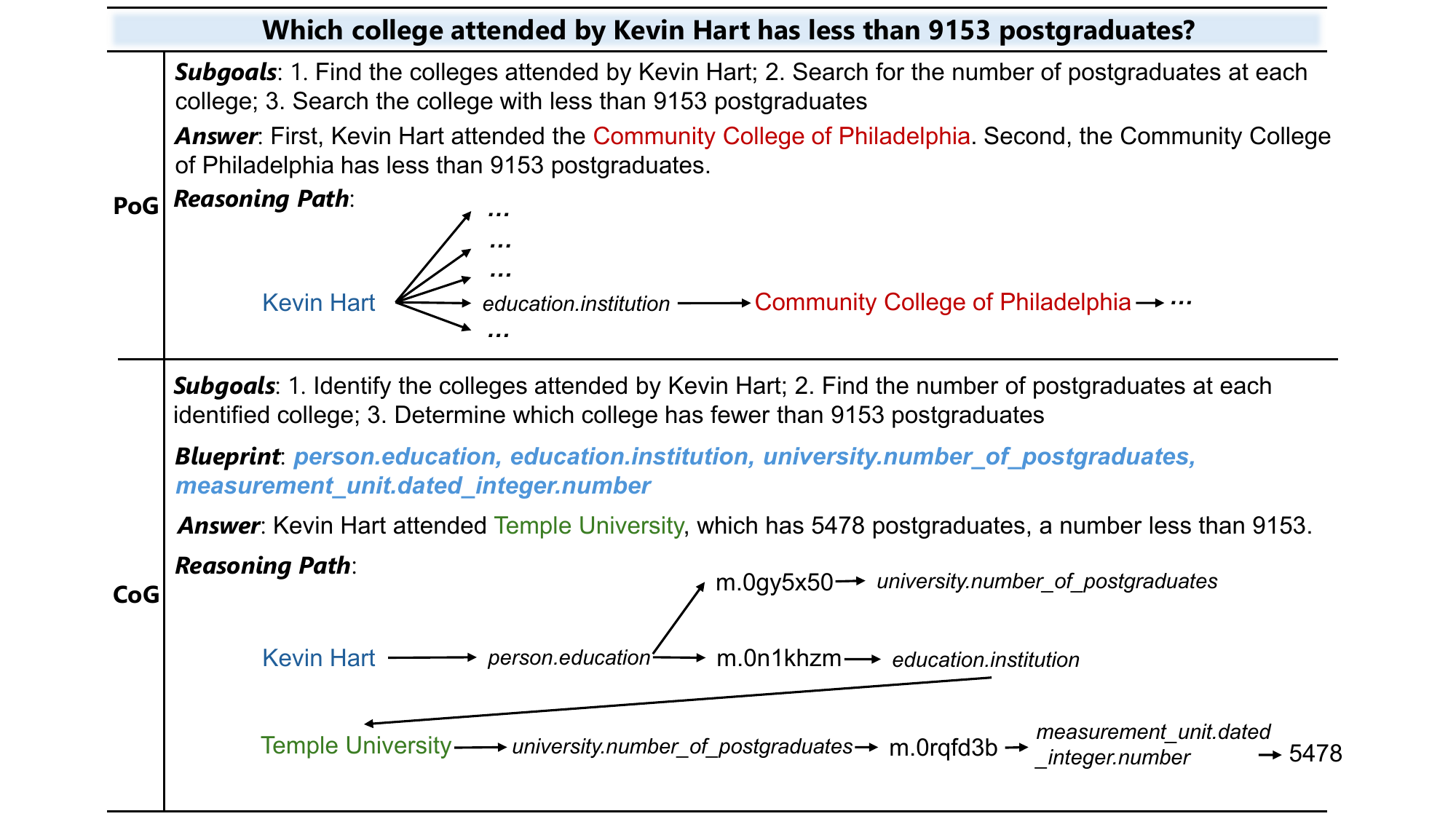}
    \caption{Case 2: Visualization of CoG's diagnostic re-routing. Unlike the stochastic stagnation observed in PoG (26 failed repetitive calls), CoG identifies structural misalignment at the High School node and utilizes System 2 reflection to re-route toward a valid reasoning branch.}
    \label{fig:case2}
\end{figure*}

\noindent\textbf{Case 2} (Figure~\ref{fig:case2}) illustrates how CoG resolves structural misalignment through diagnostic refinement, distinguishing its systematic backtracking from the stochastic retries of existing agents. When searching for the specific college attended by Kevin Hart under postgraduate constraints, PoG correctly identifies the \textit{Community College of Philadelphia} but falls into a state of cognitive rigidity. Because its self-correction mechanism is purely heuristic and lacks structural foresight, PoG enters a loop of 26 redundant calls (14,187 tokens) on the same invalid node, unable to escape the local search impasse. Conversely, CoG leverages Failure-Aware Refinement to execute a structural diagnosis upon reaching the \textit{Castlemont High School} node. Recognizing that this entity type is ontologically incompatible with the university-level slots required by the blueprint, CoG avoids blind retries. Instead, this diagnostic failure triggers a targeted re-routing back to the education hub, leading directly to \textit{Temple University}, where it extracts the specific value (5,478) and satisfies the condition. This success underscores CoG's ability to maintain reasoning momentum where baseline agents stagnate.

\section{Error Analysis: Blueprint Matching Failures}
\label{app:error_analysis}

A fundamental challenge in retrieve-then-reason paradigms is the risk of cascading error propagation: if the initial retrieval phase yields an irrelevant or structurally flawed blueprint, it risks acting as a deceptive structural prior that misguides the LLM into irreversible hallucinations. CoG proactively neutralizes this vulnerability through the closed-loop synergy of its dual-process architecture. When equipped with a mismatched blueprint, the agent is constrained from blindly forcing semantic alignment; instead, it inevitably encounters persistent topological mismatches or reasoning dead-ends during execution on the actual Knowledge Graph. Crucially, rather than culminating in a hallucinated derivation, these structural impasses serve as explicit diagnostic signals. They trigger an immediate, high-level intervention by the Failure-Aware Refinement module (System 2). Upon detecting continuous execution anomalies, System 2 systematically diagnoses the global unreliability of the active blueprint, dynamically overrides its rigid constraints, and autonomously initiates targeted backtracking to explore valid subgraphs. Consequently, a catastrophic blueprint matching failure in CoG does not equate to an erroneous conclusion. It merely incurs the bounded computational overhead of the initial misdirected exploration, while System 2 acts as an architectural safeguard that decisively severs the error propagation chain to preserve the fidelity of the final reasoning outcome.

\end{document}